\documentclass[sigconf,nonacm]{acmart}

\usepackage{aliascnt}
\usepackage{caption}
\usepackage{subcaption}
\usepackage{csquotes}
\usepackage{acro}
\usepackage{varwidth}  
\usepackage[most]{tcolorbox}
\usepackage{booktabs}      
\usepackage{tabularx}        
\usepackage{makecell}      
\usepackage{xcolor}         
\usepackage[nomessages]{fp} 
\usepackage{multirow}
\usepackage[inline]{enumitem}
\usepackage{bm}
\usepackage{wasysym}
\usepackage{array}
\usepackage{tikz}
\usetikzlibrary{arrows.meta}
\usepackage[ruled,vlined]{algorithm2e}
\usepackage{listings}
\usepackage{stfloats}
\usepackage{bbm}

\newlength{\maxlen}

\DeclareRobustCommand{\caparrow}[1]{%
  \tikz[baseline=-0.6ex] \draw[#1, line width=0.5pt] (0,0) -- (0.3,0);%
}

\DeclareRobustCommand{\pricebox}{%
  \tikz[baseline=-0.9ex] 
    \draw[draw=black, line width=0.5pt] (0,-0.15) rectangle (0.27,0.1);%
}

\newcommand{\circnum}[1]{%
\tikz[baseline=(char.base)]{
  \node[shape=circle, draw, inner sep=0.8pt] (char) {\small #1};
}}

\makeatletter
\renewcommand{\sectionautorefname}{\S\@gobble}
\renewcommand{\subsectionautorefname}{\S\@gobble}
\renewcommand{\subsubsectionautorefname}{\S\@gobble}

\makeatother

\newtcolorbox[auto counter]{prompt}[2][]{%
  colback=white,
  colframe=blue!20!white,
  coltitle=black,
  fontupper=\footnotesize,  
  fonttitle=\footnotesize, 
  float,
  floatplacement=!h,
  title=Prompt~\thetcbcounter: #2,
  #1,
}

\newtcolorbox[auto counter]{reasoning}[2][]{%
  colback=white,
  colframe=blue!20!white,
  coltitle=black,
  fontupper=\footnotesize,  
  fonttitle=\footnotesize, 
  float,
  floatplacement=!h,
  title=CoT~\thetcbcounter: #2,
  #1,
}

\setlist[itemize]{
  leftmargin=*,
  labelsep=0.5em,
  topsep=2pt,
  itemsep=0pt,
  parsep=0pt,
  partopsep=0pt
}

\newcommand{\dec}[1]{\textcolor{red}{\tiny $\blacktriangledown$#1}}

\definecolor{jsonkey}{rgb}{0.55,0.0,0.0}
\definecolor{jsonstring}{rgb}{0.0,0.5,0.0}
\definecolor{jsoncomment}{rgb}{0.5,0.5,0.5}

\lstdefinelanguage{json}{
  basicstyle=\ttfamily\small,
  numbers=none,
  showstringspaces=false,
  breaklines=true,
  frame=single,
  backgroundcolor=\color{gray!5},
  string=[s]{"}{"},
  comment=[l]{//},
  morecomment=[s]{/*}{*/},
  morestring=[b]',
  literate=
   *{0}{{{\color{blue}0}}}{1}
    {1}{{{\color{blue}1}}}{1}
    {2}{{{\color{blue}2}}}{1}
    {3}{{{\color{blue}3}}}{1}
    {4}{{{\color{blue}4}}}{1}
    {5}{{{\color{blue}5}}}{1}
    {6}{{{\color{blue}6}}}{1}
    {7}{{{\color{blue}7}}}{1}
    {8}{{{\color{blue}8}}}{1}
    {9}{{{\color{blue}9}}}{1}
}

\lstdefinestyle{jsonplain}{
  language=json,
  basicstyle=\ttfamily\scriptsize,
  numbers=none,              
  showstringspaces=false,
  breaklines=true,
  breakautoindent=false, 
  breakindent=0pt,     
  frame=single,
  backgroundcolor=\color{gray!5},
  xleftmargin=0pt,          
  framexleftmargin=0pt,
  aboveskip=4pt,
  belowskip=4pt,
  columns=fullflexible,
  keepspaces=true,
}

\newtheorem{theorem}{Theorem}

\newaliascnt{lemma}{theorem}
\newtheorem{lemma}[lemma]{Lemma}
\aliascntresetthe{lemma}

\DeclareAcronym{mas}{
    short = MAS,
    long = multi-agent system
}

\DeclareAcronym{dex}{
    short = DEX,
    long = decentralized exchange
}

\DeclareAcronym{cot}{
    short = CoT,
    long = chain-of-thought
}

\DeclareAcronym{kol}{
    short = KOL,
    long = key opinion leader
}

\DeclareAcronym{defi}{
    short = DeFi,
    long = decentralized finance
}

\DeclareAcronym{llm}{
    short = LLM,
    long = large language model
}

\DeclareAcronym{amm}{
    short = AMM,
    long  = Automatic Market Maker
}

\DeclareAcronym{nn}{
    short = NN,
    long  = neural network
}

\DeclareAcronym{xgboost}{
    short = XGBoost,
    long  = eXtreme Gradient Boosting
}

\DeclareAcronym{mlp}{
    short = MLP,
    long  = multilayer perceptron
}

\DeclareAcronym{lasso}{
    short = LASSO,
    long  = least absolute shrinkage and selection operator
}

\DeclareAcronym{auc}{
    short = AUC,
    long  = Area Under Curve
}

\DeclareAcronym{roc}{
    short = ROC,
    long  = receiver operating characteristic
}

\DeclareAcronym{aucroc}{
    short = AUC-ROC,
    long  = area under the receiver operating characteristic
curve
}

\DeclareAcronym{shap}{
  short = SHAP,
  long  = SHapley Additive exPlanations,
}

\AtBeginDocument{%
  }

\copyrightyear{2026}
\acmYear{2026}
\setcopyright{cc}
\setcctype{by}
\acmConference[WWW '26]{Proceedings of the ACM Web Conference 2026}{April 13--17, 2026}{Dubai, United Arab Emirates}
\acmBooktitle{Proceedings of the ACM Web Conference 2026 (WWW '26), April 13--17, 2026, Dubai, United Arab Emirates}
\acmPrice{}
\acmDOI{10.1145/3774904.3792635}
\acmISBN{979-8-4007-2307-0/2026/04}

\begin{document}

\title[Resisting Manipulative Bots in Meme Coin Copy Trading]{%
Resisting Manipulative Bots in Meme Coin Copy Trading:\\
A Multi-Agent Approach with Chain-of-Thought Reasoning}

\author{Yichen Luo}
\affiliation{
  \institution{UCL, Centre for Blockchain Technologies}
  \city{London}
  \country{United Kingdom}
}
\affiliation{
  \institution{The University of Hong Kong, FinTech Academy}
  \city{Hong Kong SAR}
  \country{China}
}
\email{yichen.luo.22@ucl.ac.uk}

\author{Yebo Feng}
\affiliation{
  \institution{Nanyang Technological University}
  \country{Singapore}
}
\email{yebo.feng@ntu.edu.sg}
\authornote{Yebo Feng is the corresponding author.}

\author{Jiahua Xu}
\affiliation{
 \institution{UCL, Centre for Blockchain Technologies}
 \institution{DLT Science Foundation}
 \city{London}
 \country{United Kingdom}
 }
\email{jiahua.xu@ucl.ac.uk}

\author{Yang Liu}
\affiliation{
  \institution{Nanyang Technological University}
  \country{Singapore}
}
\email{yangliu@ntu.edu.sg}

\begin{abstract}
    Copy trading has become the dominant entry strategy in meme coin markets. However, due to the market's extremely illiquid and volatile nature, the strategy exposes an exploitable attack surface: adversaries deploy manipulative bots to front-run trades, conceal positions, and fabricate sentiment, systematically extracting value from na\"ive copiers at scale. Despite its prevalence, bot-driven manipulation remains largely unexplored, and no robust defensive framework exists. We propose a manipulation-resistant copy-trading system based on a multi-agent architecture powered by a multi-modal \ac{llm} and \ac{cot} reasoning. Our approach outperforms zero-shot and most statistic-driven baselines in prediction accuracy as well as all baselines in economic performance, achieving an average copier return of 3\% per meme coin investment under realistic market frictions. Overall, our results demonstrate the effectiveness of agent-based defenses and predictability of trader profitability in adversarial meme coin markets, providing a practical foundation for robust copy trading.
\end{abstract}




\maketitle

\section{Introduction}

While manipulative tactics such as rat trading are well-trodden ruses in traditional financial markets, these old-fashioned ploys have now become bot-driven and prey on na\"ive copy traders in a new hunting ground: the meme coin market. The launch of the \$TRUMP meme coin on January 17, 2025 ignited a surge of speculation in the meme coin market, drawing millions of traders. However, many of these entrants lack prior trading experience. To lower the barrier to participation, meme coin tracking platforms such as GMGN introduced \href{https://gmgn.ai/trade/66uSvVo9?chain=sol}{\textcolor{cyan}{copy trading}}, an automated, one-click feature that allows users to replicate the trades of other wallets. In practice, copiers attempt to identify so-called smart money---wallets perceived to possess insider knowledge, trading expertise, and consistent profitability---and replicate their trading activity~\cite{Apesteguia2020CopyTrading,Shen2025Irrationality-AwareTrading}.

However, the characteristics of the meme coin market, most notably extreme illiquidity and high volatility, combined with the na\"{\i}vet\'e of copy traders, create an ideal environment for exploitation. Sophisticated adversaries deploy automated trading bots (e.g., \href{https://pumpkingbundler.com/}{\textcolor{cyan}{PumpKing}}) to systematically manipulate prices and induce predictable copy-trading behavior. In particular, some \acp{kol} featured on tracking platforms may abuse their perceived credibility by coordinating multiple bots to accumulate positions early at low prices, artificially inflate prices while concealing their true exposure, and engage in wash trading and fabricated on-chain or social activity to create an illusion of  demand~\cite{Ding2025DecomposeMarket,Mongardini2026AEcosystem,Kawai2024StrangerPlatforms}. These optics of profitability attract copiers, inflate prices, and allow manipulators to gain by using exit liquidity from copy traders. Despite the prevalence of these attacks, neither industry nor academia has established a copy-trading framework robust to manipulative bots, hence our motivation to design such a framework.

Recent advances in \acp{llm} offer a promising foundation for our design. Leveraging their strong capabilities in multi-modal understanding and reasoning~\cite{Yin2024AModels,Yang2024HarnessingBeyond,Li2023LargeSurvey,Zhang2024Vision-LanguageSurvey,Wei2022Chain-of-ThoughtModels,Liu2023ChainFeedback}, \ac{llm}-powered \acp{mas} are well suited for bot-resilient copy trading. First, the on-chain meme coin market can be encoded as a machine-readable environment for sequential copy-trading decisions, in which agents interact with multi-modal on-chain and off-chain signals while reasoning over latent factors such as market sentiment and sophisticated manipulation. This formulation aligns well with the multi-modal inference and reasoning strengths of \ac{llm}-powered agent systems. Moreover, investment decisions in traditional financial institutions are typically made by teams rather than individuals; \acp{mas} mirror this collective human cognitive decision-making process by enabling multiple specialized agents to collaborate, deliberate, and reach more robust copy-trading decisions, thereby improving resilience against manipulative bots and adversarial trading behavior. 

However, due to limited domain knowledge of meme coin markets, zero-shot \acp{llm} often fail to identify subtle manipulative patterns or reliably distinguish genuine smart money from adversarial wallets. \Ac{cot} reasoning, in contrast, injects domain knowledge into agent deliberation, enabling agents to internalize copy-trading knowledge and reason in a manner closer to experienced human copy traders.

Accordingly, we propose an \ac{llm}-powered \ac{mas} for manipulation-resistant meme coin copy trading with \ac{cot} reasoning. Our framework decomposes copy trading into three core subtasks---coin evaluation, wallet selection, and timing assessment---each handled by a specialized agent guided by few-shot \ac{cot} prompts. We evaluate our framework on historical wallet-level data from 6,000 meme coins and show that the proposed multi-agent system outperforms the zero-shot and most statistic-driven baselines in prediction accuracy. Economically, the smart-money wallets identified by our framework achieve an average return of 14\%, while the corresponding estimated copier return is 3\% per meme coin investment under realistic market frictions. Our contributions are multifold:
\begin{itemize}

\item We propose an \ac{llm}-based \ac{mas} for meme coin copy trading. To the best of our knowledge, this is the \emph{first} application of a multi-agent system to copy trading.

\item We provide a comprehensive analysis of manipulative bots in meme coin markets under copy-trading settings. We formalize manipulation heuristics, develop detection algorithms, and empirically quantify bot activity in the meme coin market.

\item We design a structured feature space tailored to meme coin markets that captures manipulative bot presence as well as traders’ historical performance and experience. Using these features and proposed prediction models, we show that trader profitability is predictable, providing a foundation for meme coin copy trading.

\end{itemize}
\section{Related Works}
We next review related work on crypto market manipulation and \ac{llm}-powered agents in blockchain to highlight the gaps our work addresses in adversarial copy-trading settings.

\textbf{Crypto Market Manipulation.} Manipulation in cryptocurrency markets has been widely documented, including pump-and-dump schemes~\cite{Xu2019TheScheme,Wu2025ProfitLearning,Dhawan2023AMarkets}, rug pull~\cite{Nguyen2023Rug-pullEngineering,Mazorra2022DoDetection,Sun2025SoK:Assessment}, wash trading~\cite{Cong2023CryptoTrading,Aloosh2024DirectTrading,Niu2024UnveilingMarkets}, metric distortion~\cite{Luo2026PiercingReappraised,Perez2020RevisitingBlockchains}, and wallet drainers~\cite{Erinle2025SoK:Wallets}. 
In the context of meme coin markets, \citet{Mongardini2026AEcosystem} and \citet{Li2025TrustMarkets} document the prevalence of systematic manipulation. \citet{Ding2025DecomposeMarket} empirically quantify the causal effects of distinct meme coin manipulation strategies on project performance and participant profitability. \citet{Xiang2025MeasuringFragility} develop a framework to quantify the fragility in the meme coin market. This line of work primarily focuses on measuring manipulation, whereas defensive mechanisms for copy trading under adversarial conditions remain underexplored.

\textbf{\ac{llm}-powered Agents in Blockchain.}
Recent advances in \ac{llm}-powered autonomous agents have begun to transform the blockchain ecosystem. Prior work has explored the use of \ac{llm}-based agents in cryptocurrency markets and \ac{defi}, particularly in quantitative trading~\cite{Luo2025LLM-PoweredManagement,Li2024CryptoTrade:Trading,Wang2024ExploringReasoning}, smart contract auditing~\cite{Wei2025AdvancedSystems}, and user intent understanding~\cite{Mao2025KnowMining}.
Building on evidence that \ac{cot} prompting substantially enhances the reasoning capabilities of \acp{llm}~\cite{Wei2022Chain-of-ThoughtModels,Xiong2025DeliberateModel,Xiong2025EnhancingAggregation}, recent studies have increasingly incorporated \ac{cot}-based reasoning to improve decision-making in financial agents~\cite{Zhong2025DetectingReasoning,Gao2025AiRacleX:Generation,Xie2025FinChain:Reasoning,Nitarach2025FinCoT:Reasoning}. However, the application of multi-agent \ac{llm} systems to adversarially robust copy trading in meme coin markets has received limited attention.
 
\section{Background}
\subsection{Solana and Decentralized Finance}
\label{subsec:solana_and_decentralized_finance}

Solana is a high-throughput permissionless blockchain that supports \ac{defi} applications via smart contracts and a block-based execution model. It features sub-second block times (approximately 400\,ms) and processes thousands of transactions per block, enabling rapid on-chain state updates and dense trading activity. At the time of writing, Solana has processed over 500 billion transactions. Leveraging its high throughput, Solana does not rely on a public mempool in which pending transactions remain visible for an extended period prior to block inclusion. Instead, transactions are forwarded directly to validators and incorporated into blocks with minimal delay, leaving little time for the general public to observe pending transactions and react to emerging trading activity. As a result, the ability to respond strategically, such as through front-running or coordination, is concentrated among well-positioned or infrastructure-level participants.

\ac{defi} activity on Solana consists of repeated interactions between wallets and the evolving on-chain state. At each block, wallets observe prices, balances, and transactions, and submit actions such as token creation, buys, sells, and transfers. State updates follow deterministic smart-contract logic~\cite{Xu2025Auto.gov:DeFi}, but these trading decisions are subject to uncertainty from transaction ordering, concurrent wallet activity, and latent coordination.

\subsection{Meme Coin Markets and Copy Trading}
\label{subsec:memecoin_markets_and_copy_trading}
\begin{figure}[tb]
  \centering
  \begin{subfigure}{0.49\linewidth}
      \centering
      \includegraphics[width=0.96\linewidth]{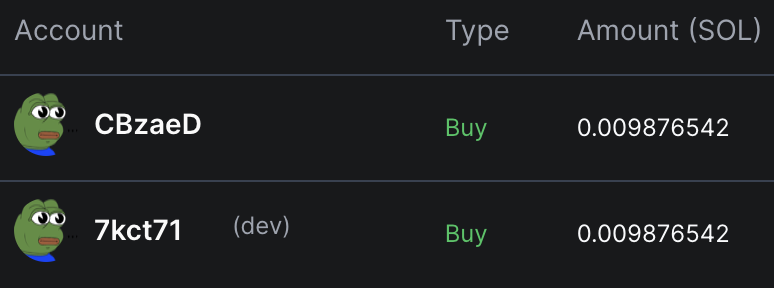}
      \caption{Transaction History.}
      \label{subfig:txn}
    \end{subfigure}%
    \hfill
    \begin{subfigure}{0.5\linewidth}
      \centering
      \includegraphics[width=\linewidth]{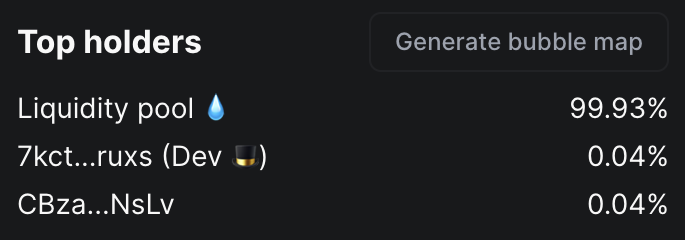}
      \caption{Holding Concentration.}
      \label{subfig:holder}
    \end{subfigure}
    \caption{Excerpts of transaction history and token holding concentration on pump.fun of meme coin \href{https://pump.fun/coin/EiJKTSMthPWQtxg1JAhbmt6t2ArW5zHCt9iT3xE8pump}{\textcolor{cyan}{Oracle}}. The label \enquote{dev} denotes the meme coin creator, whereas \enquote{liquidity pool} denotes unpurchased coins.}
  \label{fig:txn_and_holding}
\end{figure}

Meme coins are cryptocurrencies originating from internet memes \cite{Long2025BridgingEcosystem,Long2025CoinCLIP:Memecoins}. They are typically characterized by low liquidity and extreme volatility, with their prices driven by community hype, social media activity, and celebrity endorsements. Pump.fun is the largest on-chain meme coin launch platform on Solana, enabling users to rapidly create and trade meme coins. Traders purchase meme coins using SOL, Solana’s native cryptocurrency, paying a 1\% transaction fee and following a bonding curve mechanism. The bonding curve induces a strictly increasing and convex price path, so prices rise deterministically with cumulative demand, yielding diminishing tokens received per unit of SOL deposited. As a result, late entry is mechanically penalized along the curve (see \autoref{apx:bonding_curve_mechanism}). Besides the financial market, pump.fun also provides a social platform. Any user with a Solana wallet can comment on any coin and interact with other users by referencing specific comments via unique comment IDs (e.g., \enquote{\href{https://pump.fun/coin/DNnzSTbL8z4x28ji2dDmZKnYAbLf8zcmqiG9UjArpump\#p144840307}{\textcolor{cyan}{@144840307 it is wild}}}). Each trade bumps the coin’s name to the front page with a brief jiggle. As illustrated in \autoref{fig:txn_and_holding}, pump.fun publicly displays both transaction histories and token holding distributions, with the meme coin creator explicitly labeled.

In parallel, meme coin tracking platforms such as GMGN aggregate on-chain trading activity and highlight so-called \ac{kol} wallets that are ranked based on returns or number of trades, providing copy-trading tools to replicate trades from selected addresses.

\section{Modeling and Detecting Manipulations}
\begin{figure*}[!b]
  \centering
  \begin{subfigure}{0.175\linewidth}
      \centering
      \includegraphics[width=\linewidth]{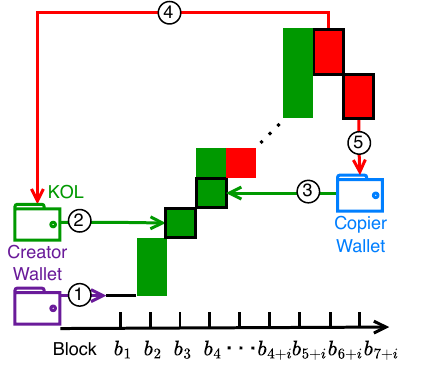}
      \caption{Benign Trading.}
      \label{subfig:candle_organic}
    \end{subfigure}%
    \hfill
  \begin{subfigure}{0.16\linewidth}
      \centering
      \includegraphics[width=0.9\linewidth]{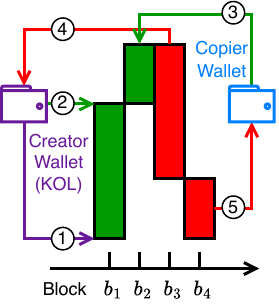}
      \caption{Na\"ive Bundle.}
      \label{subfig:candle_naive}
    \end{subfigure}%
    \hfill
    \begin{subfigure}{0.25\linewidth}
      \centering
      \includegraphics[width=0.85\linewidth]{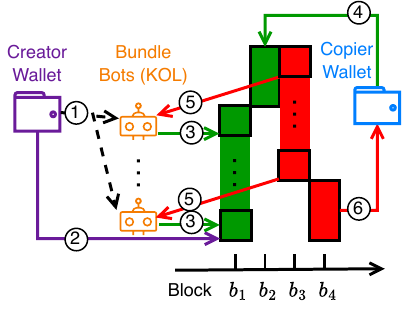}
      \caption{Bundle Bot.}
      \label{subfig:candle_launch_bundle}
    \end{subfigure}%
    \hfill
    \begin{subfigure}{0.25\linewidth}
      \centering
      \includegraphics[width=\linewidth]{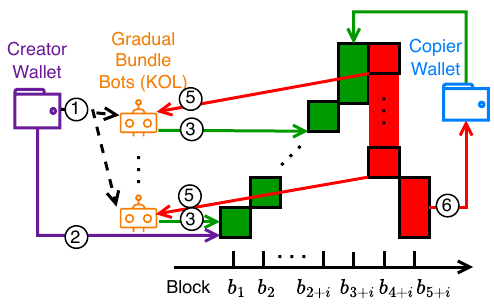}
      \caption{Gradual Bundle.}
      \label{subfig:candle_gradual_bundle}
    \end{subfigure}
    \begin{subfigure}{0.16\linewidth}
      \centering
      \includegraphics[width=\linewidth]{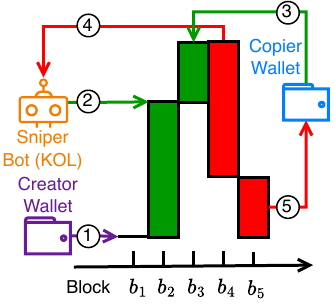}
      \caption{Sniper Bot.}
      \label{subfig:candle_sniper}
    \end{subfigure}%
    \hfill
    \caption{Evolution and heuristics of position manipulation bots in copy trading, as detailed in \autoref{subsec:manipulative_bot_behaviors}.
\caparrow{->, draw=green!40!black} denotes buys,
\caparrow{->, draw=red} denotes sells,
\caparrow{->, draw=violet!80!black} denotes coin creation, and
\caparrow{->, draw=black, dashed} denotes coin transfer.
\pricebox\ denotes price impact of each buy or sell action. The horizontal axis shows block number, and circled numbers indicate action order. A \textcolor{orange!50!white}{bot} is a wallet controlled by an automated script. For each candlestick, the bottom and top of the body indicate the price at the beginning and end of the block, respectively: a \textcolor{green!40!black}{green candle} indicates that the end-of-block price is higher than the start-of-block price, while a \textcolor{red}{red candle} indicates the opposite.}
\label{fig:heuristic_candle}
\end{figure*}
This section describes manipulations in detail. We introduce our threat model, discuss ways to carry out manipulation, propose detection metrics and algorithms, and present detection results.

\subsection{Threat Model}
We assume a strategic adversary capable of rapidly creating and controlling a large number of on-chain wallets. The adversary can coordinate these wallets to create, buy, and sell meme coins with precise timing and high execution speed, thereby ensuring transaction inclusion in targeted blocks.

The adversary’s ultimate objective is to extract economic value from copy traders. Specifically, the adversary seeks to
\begin{enumerate*}[label=(\roman*)]
    \item select a historically profitable wallet to attract victims on meme coin tracking websites or social media,
    \item acquire meme coins at very early stages, either by launching the meme coins themselves (as \emph{bundler}) or by entering immediately after meme coin creation (as \emph{sniper}),
    \item conceal true positions by distributing purchases across multiple controlled wallets,
    \item fake trading volume and social-media sentiment through wash trading and coordinated comment bots, thereby increasing copier confidence, and
    \item strategically exit positions by coordinating sales across controlled wallets, thereby further strengthening the appearance of consistent profitability and sustaining the manipulation cycle.
\end{enumerate*}

We model the interaction between the adversary and copy traders as a near zero-sum game. The adversary’s profits come directly from the losses of copy traders who follow manipulated on-chain and off-chain signals.

\subsection{Manipulative Bot Behaviors}
\label{subsec:manipulative_bot_behaviors}

To operationalize at scale, the adversary deploys automated bots that programmatically control wallets and execute transactions. We categorize these bots into two broad classes based on their primary role: \emph{position manipulation bots} and \emph{attention manipulation bots}.

\subsubsection{Position Manipulation Bots}

\emph{Position manipulation bots}, including \emph{bundle bots} and \emph{sniper bots}, exploit temporal advantages at or near meme coin creation to secure (and conceal) positions before copy traders can react. Beyond early acquisition, these bots coordinate wallet splitting and synchronized execution to obfuscate true positions and evade detection. We further illustrate the underlying mechanisms and heterogeneity of frontrunning and concealment strategies. \autoref{fig:heuristic_candle} provides evolution and heuristics of position manipulation bots in copy trading.
\autoref{subfig:candle_organic} illustrates an unmanipulated copy-trading process. The meme coin creator launches a meme coin at block $b_1$ (\circnum{1}). A \ac{kol} wallet identifies the meme coin’s potential and purchases it at block $b_3$ (\circnum{2}). A copier wallet imitates the \ac{kol}’s action and buys the meme coin at block $b_4$ (\circnum{3}). From block $b_4$ to $b_{4+i}$,  additional retail traders enter the market, trade the meme coin, and drive the price upward. The \ac{kol} sells the meme coin at a profit at block $b_{6+i}$ (\circnum{4}). The copier wallet also sells the meme coin at a profit at block $b_{7+i}$ (\circnum{5}).

While the process above reflects an unmanipulated copy-trading scenario, it assumes that the \ac{kol} acts honestly. To maximize profit, a \ac{kol} may instead employ a strategy known as a \emph{na\"ive bundle}, as illustrated in \autoref{subfig:candle_naive}. Unlike unmanipulated copy trading, the \ac{kol} in this strategy creates the meme coin and purchases it themself within the same block $b_1$ (\circnum{1}). When a copier wallet subsequently purchases the meme coin (\circnum{2}), the \ac{kol} (also the creator) immediately sells the meme coin for a profit (\circnum{3}), leaving the copiers with a loss (\circnum{4}).

However, this na\"ive strategy is easy to detect, since pump.fun and other meme coin tracking platforms publicly expose transaction histories and token holding distributions, with the meme coin creator explicitly identified (see \autoref{subsec:memecoin_markets_and_copy_trading} and \autoref{fig:txn_and_holding}). As a result, copiers can readily identify the creator’s concentrated ownership and avoid such meme coins. In response, adversaries adopt more sophisticated strategies, namely, the \emph{bundle bot} (\autoref{subfig:candle_launch_bundle}) and the \emph{gradual bundle bot} (\autoref{subfig:candle_gradual_bundle}), to decouple meme coin creation from accumulation. In these strategies, the \ac{kol} first creates and funds multiple fresh wallets (\circnum{1}), launches the meme coin (\circnum{2}), and then uses automated scripts to coordinate purchases across these wallets, either within a single block (\autoref{subfig:candle_launch_bundle}) or gradually over multiple blocks (\autoref{subfig:candle_gradual_bundle}) (\circnum{3}). Because the purchasing wallets appear independent, copiers are less likely to associate them with the meme coin creator and are more inclined to follow the observed buying activity (\circnum{4}). Finally, once sufficient copy-trading demand has accumulated, the \ac{kol} coordinates these wallets to exit their positions and dump the meme coin (\circnum{5}), leaving the copier with a loss (\circnum{6}).

In addition to bundle-based strategies, adversaries may employ a \emph{sniper bot} to exploit copy trading without participating in meme coin creation, as shown in \autoref{subfig:candle_sniper}. A sniper bot monitors newly launched meme coins and identifies those that have not yet been bundled by their creators. Upon detecting such a meme coin (\circnum{1}), the bot executes purchases within a very short time window following meme coin creation, typically within the first one to five blocks (approximately 0.4 to 2 seconds on Solana, see \autoref{subsec:solana_and_decentralized_finance}) (\circnum{2}). This early purchase subsequently attracts copier wallets, which replicate the observed buying activity and enter the market (\circnum{3}), driving up the meme coin price. The sniper bot then exits its position for a profit (\circnum{4}), leaving copier wallets exposed to losses (\circnum{5}).

\subsubsection{Attention Manipulation Bots}
\begin{figure}[tb]
  \centering
  \begin{subfigure}{\linewidth}
      \centering
      \includegraphics[width=0.8\linewidth]{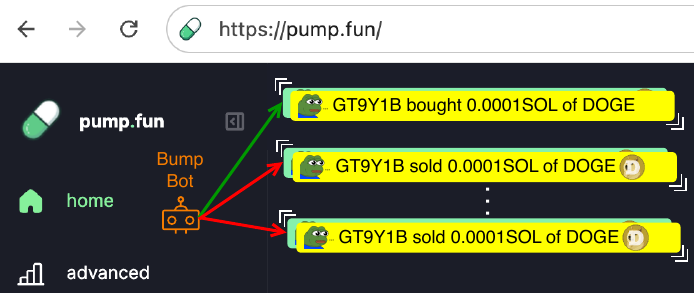}
      \caption{Bump Bot.}
      \label{subfig:bump_bot}
    \end{subfigure}
    \begin{subfigure}{\linewidth}
      \centering
      \includegraphics[width=0.8\linewidth]{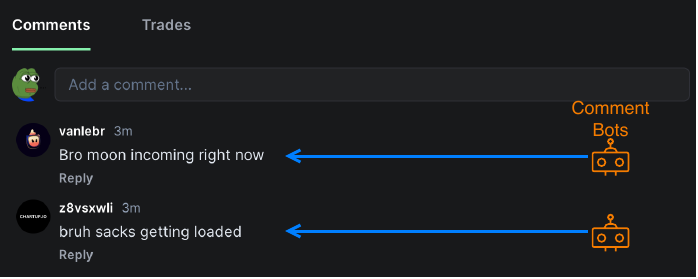}
      \caption{Comment Bot.}
      \label{subfig:comment_bot}
    \end{subfigure}
    \caption{Heuristics of attention manipulation bots.
    \caparrow{->, draw=green!70!black} denotes buys,
    \caparrow{->, draw=red!75!black} denotes sells,
    and \caparrow{->, draw=blue!75!black} denotes comments.
A \textcolor{orange!50!white}{bot} represents a wallet controlled by an automated script.
}
  \label{fig:heuristic_attention}
\end{figure}

\emph{Attention manipulation bots} generate artificial trading activity and coordinated social signals to capture and amplify attention around specific meme coins. By faking demand and legitimacy, these bots steer copy traders’ attention toward targeted meme coins, reinforce their confidence in following observed trading behavior, and attract additional uninformed retail participants. We further decompose attention manipulation bots into two representative categories: \emph{bump bots} and \emph{comment bots}. \autoref{fig:heuristic_attention} illustrates the heuristics of these bots, highlighting how on-chain activity and off-chain signals jointly amplify attention.

Bump bots exploit the bumping mechanism in pump.fun by repeatedly buying and selling identical amounts of a meme coin, thereby inflating short-term trading activity and visibility while maintaining minimal net exposure, as illustrated in \autoref{subfig:bump_bot}. Comment bots, in contrast, consist of wallets that automatically generate coordinated hype comments to simulate community engagement and amplify perceived interest, as shown in \autoref{subfig:comment_bot}\footnote{
``Moon'' is crypto jargon for a sharp price increase, while ``sacks'' is crypto jargon for accumulated capital.}.

\subsection{Detection Metrics and Algorithms}
\label{subsec:proposed_metrics_and_algorithms}
We next explain how we detect manipulative bots and the results of the detection.

\subsubsection{Bundle Bot}

Based on the heuristic illustrated in \autoref{subfig:candle_launch_bundle}, a key characteristic of bundle bots is that, within the meme coin creation block, the creator wallet not only deploys the token but is immediately followed by near-simultaneous purchases from additional controlled wallets. Given Solana’s lack of a public mempool and its very short block interval (see \autoref{subsec:solana_and_decentralized_finance}), such same-block coordination is unlikely to be achieved by independent frontrunners and instead indicates pre-coordinated control across wallets. Accordingly, we detect bundle bots by checking whether non-creator wallets execute purchases within the meme coin creation block, as formalized in \autoref{alg:bundle_bot}.

\subsubsection{Sniper Bot}

Similarly, sniper bots exhibit salient temporal characteristics, as illustrated in \autoref{subfig:candle_sniper}. They typically execute purchases within the first one to five blocks after meme coin creation, a timescale well below human reaction time. We present the sniper bot detection algorithm in \autoref{alg:sniper_bot}.

\subsubsection{Bump Bot}
Unlike bundle bots and sniper bots, bump bots do not exhibit salient temporal characteristics. Instead, their behavior is characterized by repetitive trading patterns. As illustrated in \autoref{subfig:bump_bot}, bump bots repeatedly buy and sell identical amounts of a meme coin. We define a \emph{flip} as a pair of consecutive transactions with opposite directions and identical meme coin quantities to capture this repetition. Because bump bots are controlled by automated scripts with limited strategic complexity, their net positions are expected to remain largely unchanged over time, aside from minor variations due to normal market frictions. Accordingly, our metric rewards repetitive flip behavior while penalizing significant net position changes. Specifically, for meme coin $i$ and wallet $j$, we define the bump bot score as the ratio of the number of flips to the net position change:\begin{equation}
    \small
    \label{eq:bump_score}
    \alpha_{i,j} = \frac{F_{i,j}}{\Delta P_{i,j} + \epsilon},
\end{equation}where $F_{i,j}$ is the number of flips, $\Delta P_{i,j}$ is the net position change (in meme coin units), and $\epsilon$ is a small constant to avoid a division-by-zero error (default $\epsilon{=}1$). We classify wallet $j$ as a bump bot for meme coin $i$ if $\alpha_{i,j} \ge \xi$, where $\xi$ is an empirical threshold (default $\xi{=}50$).The corresponding detection procedure is given in \autoref{alg:bump_bot}.
\begin{algorithm}[tb]
\footnotesize
\caption{Bundle Bot Detection.}
\label{alg:bundle_bot}
\KwIn{Meme coin address $A$, on-chain transactions $\mathcal{T}$ for $A$}
\KwOut{$\mathbbm{1}_{\text{Bundle Bot}} \in \{0,1\}$}

Extract creator wallet $w_c$ and launch block $b_{\text{launch}}$ for $A$\;
$\mathcal{T}_{\text{launch}} \leftarrow \{ tx \in \mathcal{T} \mid tx.\text{block} = b_{\text{launch}} \}$\;

\If{$\exists\, tx \in \mathcal{T}_{\text{launch}} \text{ s.t. } tx.\text{type}=\text{buy} \land tx.\text{trader}\neq w_c$}{
    \Return $1$\tcp*{simultaneous non-creator buys in launch block}
}
\Return $0$\;
\end{algorithm}
\begin{algorithm}[tb]
\footnotesize
\caption{Sniper Bot Detection.}
\label{alg:sniper_bot}
\KwIn{Meme coin address $A$, on-chain transactions $\mathcal{T}$ for $A$, block window $K$ (default $K{=}5$)}
\KwOut{$\mathbbm{1}_{\text{Sniper Bot}} \in \{0,1\}$}

Extract launch block $b_{\text{launch}}$ and creator wallet $w_c$ for $A$\;
$\mathcal{T}_{\text{early}} \leftarrow \{ tx \in \mathcal{T} \mid b_{\text{launch}} < tx.\text{block} \le b_{\text{launch}}{+}K \}$\;

\If{$\exists\, tx \in \mathcal{T}_{\text{early}} \text{ s.t. } tx.\text{type}=\text{buy} \land tx.\text{trader}\neq w_c$}{
    \Return $1$\tcp*{non-creator buy within first $K$ blocks}
}
\Return $0$\;
\end{algorithm}

\begin{algorithm}[tb]
\footnotesize
\caption{Bump Bot Detection.}
\label{alg:bump_bot}
\KwIn{Meme coin address $A$, on-chain transactions $\mathcal{T}$ for $A$, threshold $\xi$ (default $\xi = 50$), small constant $\epsilon$ (default $\epsilon = 1$)}
\KwOut{$\mathbbm{1}_{\text{Bump Bot}} \in \{0,1\}$}

Group $\mathcal{T}$ by trader wallet to obtain $\{\mathcal{T}_w\}$\;

\ForEach{wallet $w$ with transaction list $\mathcal{T}_w$}{
    Sort $\mathcal{T}_w$ by block height and index within block\;
    $F \leftarrow 0$\tcp*{number of flips}
    $\Delta P \leftarrow 0$\tcp*{net position change}
    
    \For{$i \leftarrow 1$ \KwTo $|\mathcal{T}_w|-1$}{
        $\Delta P \leftarrow \Delta P + \mathcal{T}_w[i]$\;
        \If{$\mathcal{T}_w[i] + \mathcal{T}_w[i+1] = 0$}{
            $F \leftarrow F + 1$\tcp*{a flip}
        }
    }

    $\Delta P \leftarrow |\Delta P|$\;
    
    \If{$\dfrac{F}{\Delta P + \epsilon} \ge \xi$}{
        \Return $1$\tcp*{wallet $w$ is classified as a bump bot}
    }
}
\Return $0$\;
\end{algorithm}

\subsubsection{Comment Bot}
\label{subsubsec:comment_bot}
The most prominent characteristic of comment bots is their use of contextless, hype-driven, and slogan-like comments (e.g., ``Bro moon incoming right now''), as illustrated in \autoref{subfig:comment_bot}. In addition, bot-generated comments typically do not reference other wallets (see \autoref{subsec:memecoin_markets_and_copy_trading}). Such patterns are well suited to detection by \ac{llm}s. Accordingly, we manually curate a set of human-generated and bot-generated comments as few-shot examples and provide them in the \ac{llm} context using Prompt \autoref{pmt:comment_bot} in \autoref{apx:comment_bot_detection_prompt}. We then use the \ac{llm} to classify comments associated with each meme coin. If more than one comment is classified as bot-generated, we label the meme coin as exhibiting comment bot activity.

\subsubsection{Gradual Bundle}
\label{subsubsec:gradual_bundle}
Gradual bundles represent a more sophisticated variant of bundle bots. As illustrated in \autoref{subfig:candle_gradual_bundle}, they decouple token creation from purchases over time and may involve multi-layer funding networks. Combined with the scale of the Solana blockchain (see \autoref{subsec:solana_and_decentralized_finance}), these properties obscure temporal signals and attribution, making gradual bundles much harder to detect accurately. In practice, experienced meme coin traders may rely on candlestick patterns to infer and avoid such coins, as the candlestick dynamics of gradual bundles differ systematically from those of non-manipulative tokens: the former tend to exhibit mechanically regular patterns, whereas the latter display more organic patterns.

\subsection{Detection Results}
\label{subsec:detection_results}
\begin{figure}[tb]
  \centering
  \begin{subfigure}{0.48\linewidth}
      \centering
        \tikz\node[inner sep=0pt,
           label=west:\rotatebox{90}{\footnotesize\textbf{Position Manipulation}}]{\includegraphics[width=\linewidth]{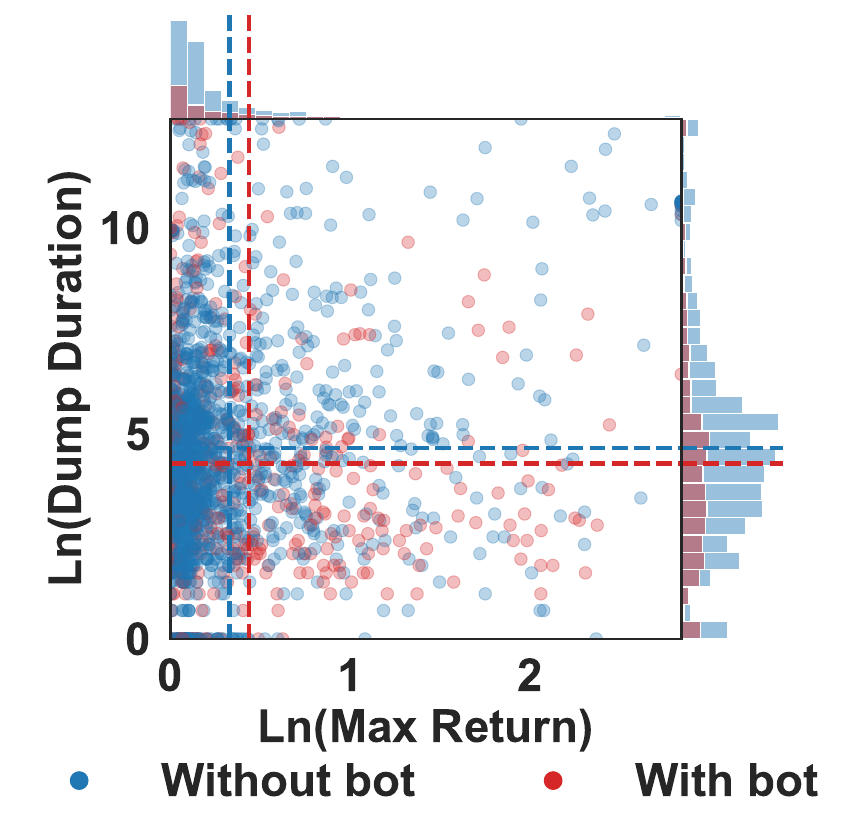}};
      \caption{Bundle Bot.}
      \label{subfig:launch_bundle_hist}
    \end{subfigure}%
    \hfill
    \begin{subfigure}{0.48\linewidth}
      \centering
      \includegraphics[width=\linewidth]{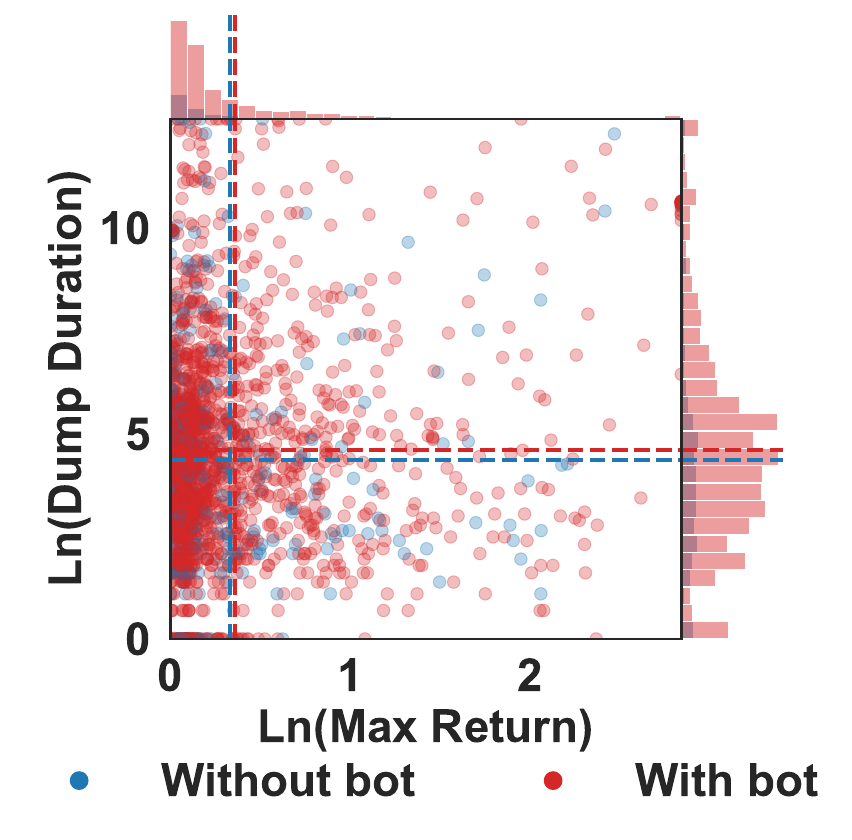}
      \caption{Sniper Bot.}
      \label{subfig:sniper_bot_hist}
    \end{subfigure}
    \vfill
    \begin{subfigure}{0.48\linewidth}
      \centering
      \tikz\node[inner sep=0pt,
           label=west:\rotatebox{90}{\footnotesize\textbf{Attention Manipulation}}]{\includegraphics[width=\linewidth]{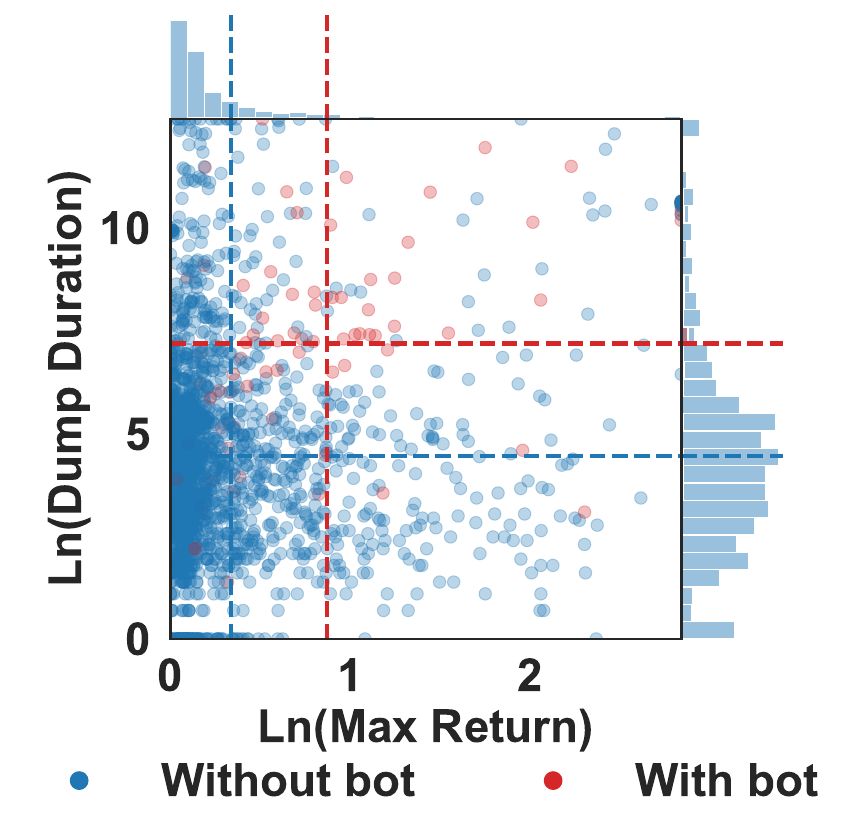}};
      \caption{Bump Bot.}
      \label{subfig:volume_bot_hist}
    \end{subfigure}%
    \hfill
    \begin{subfigure}{0.48\linewidth}
      \centering
        \includegraphics[width=\linewidth]{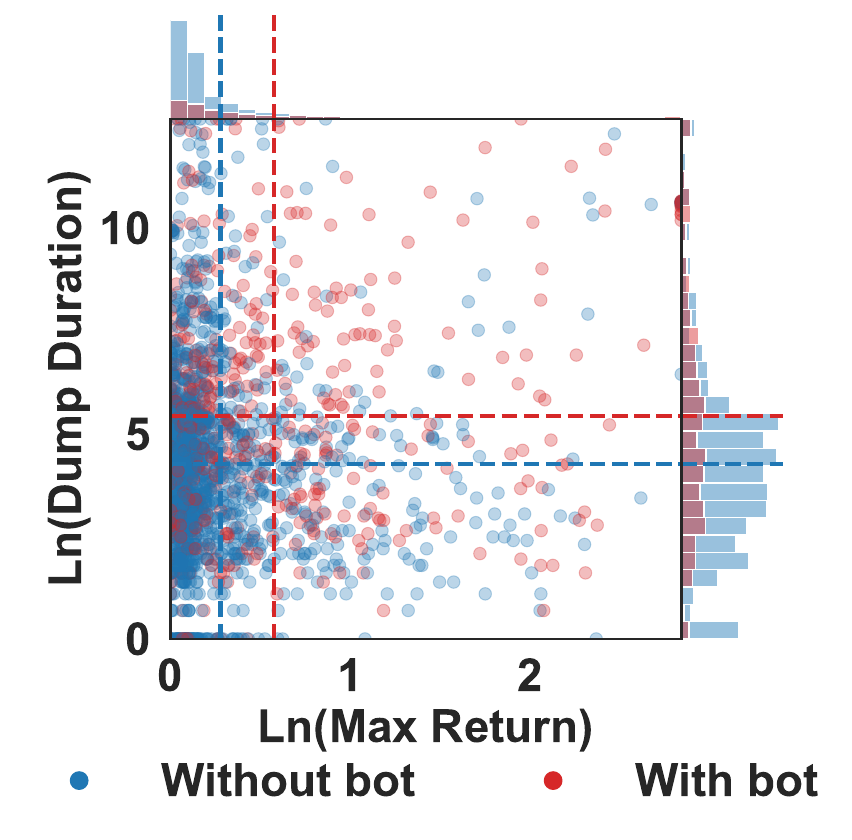}
      \caption{Comment Bot.}
      \label{subfig:bot_comment_num_hist}
    \end{subfigure}
  \caption{Distribution of performance metrics for meme coins with and without specific bots, grouped into position manipulation (first row) and attention manipulation bots (second row). The dashed line denotes the mean of the distribution.}
  \label{fig:bot_dis}
\end{figure}

For on-chain data, we use account-level transaction and transfer records obtained from Flipside, a blockchain data provider that maintains archive nodes with fully indexed, historical access to the Solana ledger. We additionally collect off-chain comment data from pump.fun. Our final dataset comprises 6{,}000 meme coin projects, including all associated creators and traders, along with their complete historical trading records.

\autoref{fig:bot_dis} presents the distributions of meme coin projects with and without four types of bots, grouped into position manipulation and attention manipulation bots, and their heterogeneous effects on two performance metrics: maximum log return and log dump duration. The maximum log return is defined as the peak price appreciation during the lift phase, $Ln(\textit{Max Return}) = \ln\!\left(P_{\max} / P_{\text{launch}}\right)$, where $P_{\max}$ and $P_{\text{launch}}$ denote the peak and launch prices, respectively. The $Ln(\textit{Dump Duration})$ is defined as the natural logarithm of seconds from the peak price until liquidity declines to 10\% of its peak level. For position manipulation bots, the performance distributions with and without bots largely overlap, indicating limited influence on either return or duration. Bundle bots appear in roughly one quarter of projects and are weakly associated with lower returns and shorter dump durations (\autoref{subfig:launch_bundle_hist}), while sniper bots are widespread but show minimal observable performance effect (\autoref{subfig:sniper_bot_hist}). This pattern suggests that the primary role of position manipulation bots is not to alter overall project performance, but to redistribute profits within the participant pool, shifting gains from slower retail traders to early or better-positioned adversaries. In contrast, attention manipulation bots, namely bump bots (\autoref{subfig:volume_bot_hist}) and comment bots (\autoref{subfig:bot_comment_num_hist}), are associated with improved performance metrics, although bump bots occur less frequently than comment bots, likely due to their higher operational cost.
\begin{table*}[!b]
    \centering
    \footnotesize
    \caption{Feature descriptions, inclusion, and conditions, as described in \autoref{subsubsec:feature_selection}.}
    \begin{tabularx}{\linewidth}{llcccl}
\toprule
\multicolumn{1}{l}{\multirow{2}{*}{\textbf{\makecell{Feature}}}} & \multicolumn{1}{l}{\multirow{2}{*}{\textbf{\makecell{Description}}}} & \multicolumn{3}{c}{\textbf{Feature Inclusion}} &
\multicolumn{1}{c}{\multirow{2}{*}{\textbf{\makecell{MAS\\Condition}}}}\\
\cmidrule{3-5}
& & \textbf{\makecell{GMGN}} & \textbf{\makecell{Statistic}} & \textbf{\makecell{MAS}} \\
\midrule

Return ($[x]$) & The average return across the previous $[x]$ participated meme coins.$^\dagger$ & \CIRCLE & \CIRCLE & \CIRCLE & $>0$\\

Number of Trades & The total number of trades executed across all participated meme coins. & \CIRCLE & \CIRCLE & \CIRCLE & $>25\%$ Training Set\\

Return Standard Deviation & The standard deviation of returns across all participated meme coins. & \Circle & \CIRCLE & \CIRCLE & $<1$\\

$t$-stat & The $t$-statistic of the trader's mean return across all participated meme coins. & \Circle & \CIRCLE & \CIRCLE & $>1.645$\\

Time Since Last Trade & The seconds elapsed since the most recent trade before the current meme coin. & \Circle & \CIRCLE & \CIRCLE & $<75\%$ Training Set\\

Time Since First Trade & The seconds elapsed since the trader's very first trade. & \Circle & \CIRCLE & \CIRCLE & $>25\%$ Training Set\\

Time Since Launch & The seconds elapsed since the meme coin's launch. & \Circle & \CIRCLE & \CIRCLE & $<75\%$ Training Set\\

Trader Purchase Price & The execution dollar price at which the trader purchases the meme coin.      & \Circle & \CIRCLE & \CIRCLE & $>25\%$ Training Set\\

Trader Purchase Amount & The total dollar value of the trader's meme coin purchase.               & \Circle & \CIRCLE & \CIRCLE & $>25\%$ Training Set\\

Trader Purchase Quantity & The number of meme coin units purchased by the trader in the transaction. & \Circle & \CIRCLE & \CIRCLE & $>25\%$ Training Set\\

$[y]$ Bot & Dummy equal to 1 if $[y]$ bot is detected at the trader's first trade, and 0 otherwise.$^\ddagger$ & \Circle & \CIRCLE & \CIRCLE & $\mathbbm{1}_{\text{Bump Bot}} =0$ \\

Candlestick & The candlestick chart of the meme coin when the trader makes the first trade. & \Circle & \Circle & \CIRCLE & No Mechanicality \\

Comments & The comments of the meme coin when the trader makes the first trade. & \Circle & \Circle & \CIRCLE & Organically Positive \\

\bottomrule
\multicolumn{5}{l}{$^\dagger x \in \{\text{all}, \text{1st}, \text{1st-5th}, \text{6th-10th}, \text{11th-15th}\}$.}\\
\multicolumn{5}{l}{$^\ddagger y \in \{\text{Bundle, Sniper, Bump, Comment}\}$.}
\end{tabularx}

    \label{tab:feature_des}
\end{table*}
\section{Bot-Resilient Copy Trading}
\label{sec:mitigation}
Based on the analysis in the previous section, we find that manipulative bots distort market efficiency, which in turn affects copy traders’ profitability. This motivates the design of bot-resilient copy-trading frameworks that help copy traders follow profitable wallets in a given meme coin while avoiding harmful bot interactions and, where appropriate, leveraging attention manipulation bots.

We therefore focus on a fundamental profitability prediction task: given a wallet $i$ that initiates its first position in meme coin $j$, we aim to predict whether the position of the wallet will ultimately be profitable (\texttt{TRUE}) or not (\texttt{FALSE}). Formally, we model this prediction as $\textit{profitable}_{i,j} = \mathcal{M}\!\left(\textit{feature}_1, \textit{feature}_2, \ldots \right),$ where $\mathcal{M}(\cdot)$ denotes our proposed \textbf{statistic-driven} and \textbf{\ac{mas}-based} models.

\subsection{Settings}
\subsubsection{Sample Specification}

We first describe the construction of our training samples. Using the wallet-level data described in \autoref{subsec:detection_results}, we chronologically order each trader’s first transaction within a given meme coin to prevent \textbf{information leakage}. Based on this temporal ordering, we partition the data into training, validation, and test sets using a 70\%–15\%–15\% split. Finally, we define the ground-truth label as the sign of the wallet’s realized profit in the corresponding meme coin.

\subsubsection{Feature Selection}
\label{subsubsec:feature_selection}
We next describe the features selected in our models. A key assumption underlying copy trading is that selected \ac{kol} wallets possess superior trading expertise or informational advantages, enabling them to achieve consistent profitability. Accordingly, we place particular emphasis on features capturing wallet experience (e.g., number of trades), as well as the magnitude (e.g., average return) and stability (e.g., return standard deviation, $t$-statistic) of historical profitability across long- (11th–15th previously traded coins), medium- (6th–10th), and short-term (1st–5th) horizons. A trader’s market entry timing is another important factor, which we capture using features such as purchase price, traded quantity, and dollar amount. In addition, we include four dummy variables indicating the presence of manipulative bots, as identified by the detection methods described in \autoref{subsec:proposed_metrics_and_algorithms}, to explicitly account for the influence of bot activity on wallet profitability. 

However, these features alone are insufficient to rule out the gradual bundle behavior described in \autoref{subsubsec:gradual_bundle} or to capture the sentiment conveyed by human-generated comments, given that meme coin valuations are largely driven by hype. Therefore, we additionally incorporate candlestick charts and comments observed at the time when the wallet trades meme coin into our \ac{mas}-based model. \autoref{tab:feature_des} explains the features included in the prediction model.

\subsubsection{Model}
\label{subsubsec:model}
\begin{figure}[tb]
  \centering
  \begin{minipage}{0.48\linewidth}
    \centering
    \begin{minipage}{0.90\linewidth}
    \centering
    \begin{lstlisting}[
language=json,
basicstyle=\ttfamily\tiny,
numbers=none,              
showstringspaces=false,
breaklines=true,
breakautoindent=false, 
breakindent=0pt,     
frame=single,
backgroundcolor=\color{gray!5},
xleftmargin=0pt,          
framexleftmargin=0pt,
aboveskip=4pt,
belowskip=4pt,
columns=fullflexible,
keepspaces=true,
]
(i) {"Bundle Bot":"Required check: Bundle Bot must be False."}

(ii) {"Candlestick":"Required check: Must not have mechanical uptrend."}

(iii) {"Statistical Significance":"Required check: t-statistic > 1.645"}

(iv) {"Trader Purchase Amount":"Required purchase amount > {25% Training Set}"}
\end{lstlisting}
    \subcaption{\ac{mas} Conditions.}
    \label{subfig:decision_boundaries}
    \end{minipage}
    \begin{subfigure}{\linewidth}
      \centering
      \includegraphics[width=\linewidth]{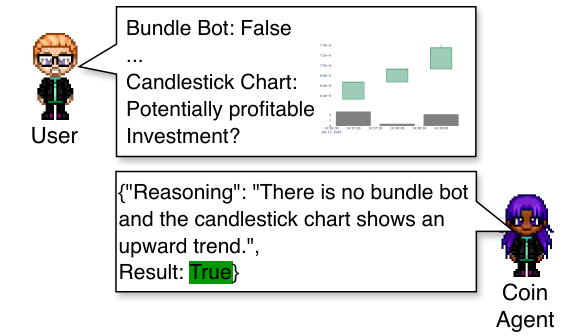}
      \subcaption{Zero-shot, without CoT.}
      \label{subfig:without_cot}
    \end{subfigure}
  \end{minipage}
  \hfill
  \begin{minipage}{0.48\linewidth}
    \centering
    \begin{subfigure}{\linewidth}
      \centering
      \includegraphics[width=\linewidth]{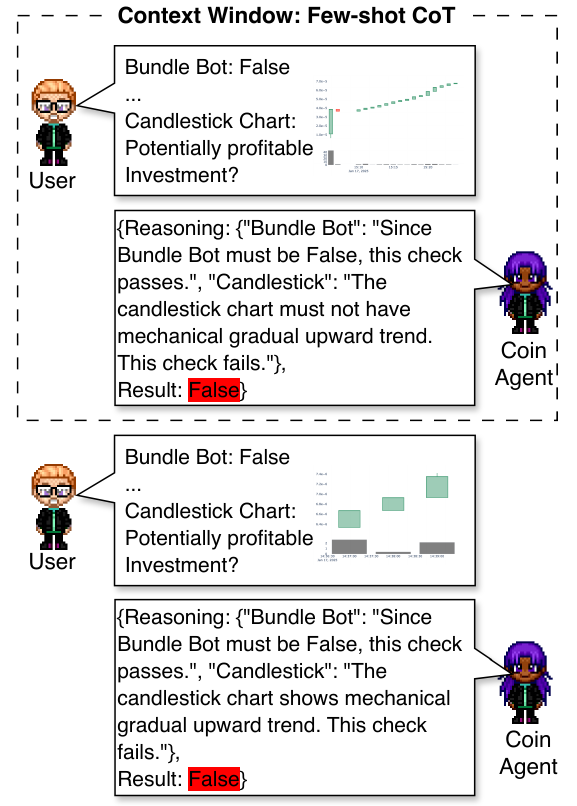}
      \subcaption{With Few-shot CoT.}
      \label{subfig:with_cot}
    \end{subfigure}
  \end{minipage}

  \caption{Comparison of model performance with and without \ac{cot}.}
  \label{fig:cot}
\end{figure}

We then describe the proposed \textbf{statistic-driven} and \textbf{\ac{mas}-based} models.

\textbf{Statistic-driven Models.} We select one representative model from each major class of machine learning methods: a linear model (LASSO), a tree-based model (XGBoost), and a \ac{nn}. Hyperparameters are tuned using the training and validation sets, and final performance is evaluated on the test set. Details of preprocessing and hyperparameter tuning are provided in \autoref{apx:hyperparameters}.

\textbf{\ac{mas}-based Models.} We introduce an \ac{llm}-powered \ac{mas}-based model for bot-resilient copy trading. The framework consists of three specialized \ac{llm} agents with \ac{cot} reasoning: a \emph{wallet agent}, a \emph{coin agent}, and a \emph{timing agent}. We employ GPT-4o as the base model for all agents\footnote{The model’s training data extends only up to October~2023, which predates all meme coin events evaluated in this study and thus precludes temporal \textbf{information leakage}.}.

\autoref{fig:cot} illustrates how domain-specific knowledge is injected into each agent via few-shot \ac{cot} prompting. For each agent, we first decide the \ac{mas} conditions defined for each feature based on~\cite{Ding2025DecomposeMarket} and annotations from professional traders~\footnote{The \ac{mas} condition design is also broadly aligned with the direction of feature contributions revealed by the \acs{shap} analysis of the statistic-driven models in \autoref{tab:feature_importance}.}. As shown in \autoref{subfig:decision_boundaries} and \autoref{tab:feature_des}, we consider four types of conditions to make the prediction \texttt{TRUE}:
\begin{enumerate*}[label=(\roman*)]
\item \textbf{Boolean} conditions enforcing binary constraints (e.g., the absence of bundle-bot behavior);
\item \textbf{Visual or language-based} conditions based on candlestick pattern or textual sentiment cues;
\item \textbf{Statistical} conditions imposing significance constraints (e.g., requiring $t$-statistics to exceed a threshold) or consistent directional effects; and
\item \textbf{Continuous} conditions requiring feature values to fall within specific percentile ranges computed from the training set.
\end{enumerate*} We then select a small set of representative examples. Next, we verify whether these examples satisfy the conditions. From these, we retain a subset of profitable and unprofitable cases that satisfy all conditions and manually annotate them to construct agent-specific \acp{cot}. These few-shot \ac{cot} exemplars are subsequently embedded into the prompts of the corresponding agents.

\begin{figure}[tb]
    \centering
    \includegraphics[width=0.9\linewidth]{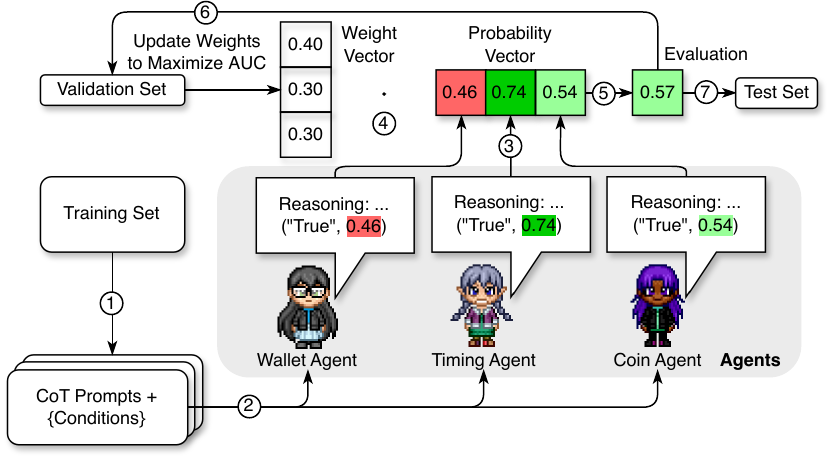}
    \caption{The architecture of our \ac{mas}-based copy-trading framework with \ac{cot} reasoning. Circled numbers indicate action order.}
    \label{fig:flow}
\end{figure}

The \ac{mas} architecture is illustrated in \autoref{fig:flow}. The step-by-step workflow of the \ac{mas} is as follows:
\circnum{1} We extract conditions for all features and incorporate these conditions into the construction of agent-specific \acp{cot}.
\circnum{2} Each agent independently performs reasoning and produces a binary \texttt{TRUE}/\texttt{FALSE} decision together with the token logits. From each output, we compute the linear probability of the \texttt{TRUE} token and interpret it as a confidence score, following established confidence-based prompting techniques for \acp{llm}~\cite{Luo2025LLM-PoweredManagement,Geng2024AModels,Kapoor2024LargeKnow,Khanmohammadi2025CalibratingStability}. The three confidence scores form a probability vector.
\circnum{3} We initialize a weight vector uniformly across agents.
\circnum{4}-\circnum{5} The weight vector is optimized on the validation set to maximize the \ac{auc} of the aggregated prediction.
\circnum{6} The final prediction is obtained as a weighted combination of agent confidence scores and is evaluated on the test set. Detailed few-shot \acp{cot} exemplars and prompts are provided in \autoref{apx:json_cot_pmt}.

To assess the efficacy of \ac{cot} reasoning, we additionally train a zero-shot \ac{mas} baseline with all other components held constant, as illustrated in \autoref{subfig:without_cot}.

\subsection{Prediction Accuracy}
\label{subsec:prediction_accuracy}
\begin{figure}[tb]
  \centering
  \begin{subfigure}{0.5\linewidth}
      \centering
      \includegraphics[width=\linewidth]{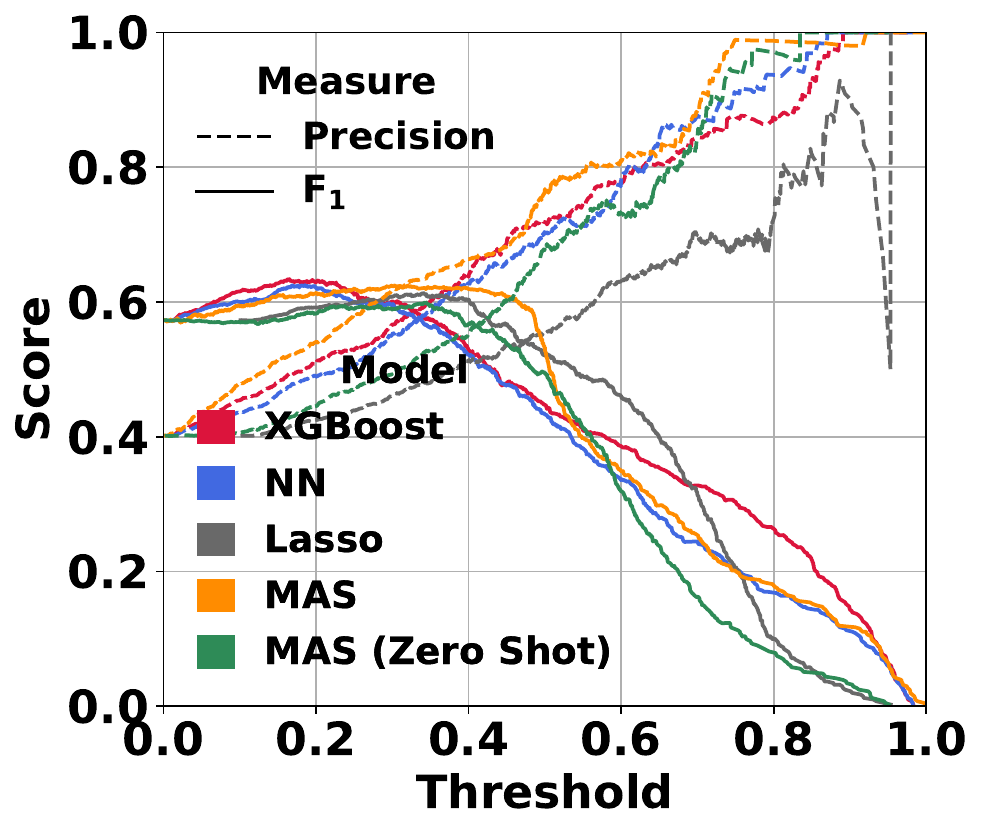}
      \subcaption{Precision and $F_1$ Scores.}
      \label{subfig:prec_f1_vs_threshold}
  \end{subfigure}%
  \hfill
  \begin{subfigure}{0.5\linewidth}
      \centering
      \includegraphics[width=\linewidth]{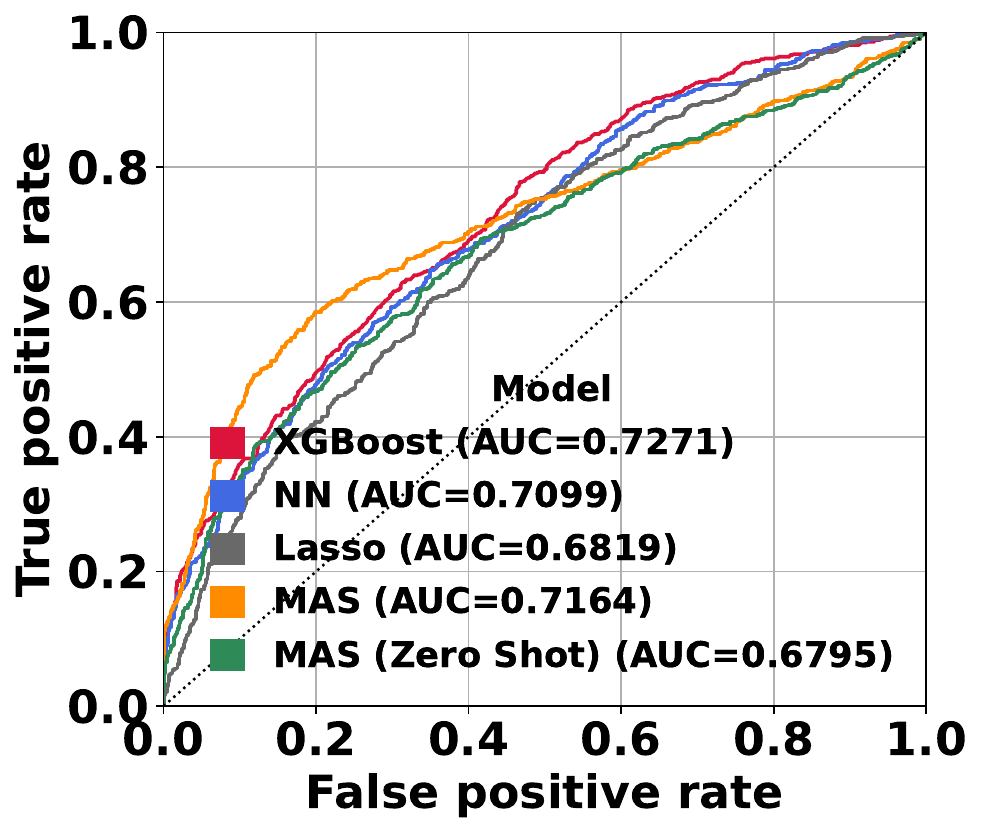}
      \subcaption{ROC Curves and AUC.}
      \label{subfig:roc_curves}
  \end{subfigure}
  \caption{Test set performance comparison between statistic-driven and \ac{mas}-based models.}
  \label{fig:prediction_accuracy}
\end{figure}

Both statistic-driven and \ac{mas}-based models output a continuous likelihood score in $[0,1]$ indicating whether a wallet is expected to be profitable. We obtain binary \texttt{TRUE}/\texttt{FALSE} predictions by applying a decision threshold. \autoref{fig:prediction_accuracy} reports out-of-sample performance on the test set as the threshold varies, using Precision, $F_1$ score, and \ac{auc} as evaluation metrics.

\autoref{subfig:prec_f1_vs_threshold} shows Precision and $F_1$ scores across thresholds. Among statistic-driven models, XGBoost consistently dominates LASSO and performs comparably to the \ac{nn}, particularly at moderate-to-high thresholds. LASSO exhibits unstable Precision and rapidly degrading $F_1$ scores as the threshold increases, reflecting its limited discriminative capacity in this setting. For \ac{mas}-based approaches, incorporating few-shot \ac{cot} reasoning leads to higher Precision and $F_1$ scores than the zero-shot \ac{mas} baseline across nearly all thresholds, indicating better calibration of confidence scores.

\autoref{subfig:roc_curves} presents \ac{roc} curves and corresponding \ac{auc} values. XGBoost achieves the highest \ac{auc} overall, followed closely by the \ac{mas} with \ac{cot} reasoning. The \ac{nn} and LASSO exhibit lower \ac{auc} values, while the zero-shot \ac{mas} performs worst, highlighting the importance of structured reasoning and agent specialization for robust ranking performance. In addition, we also present the feature importance in \autoref{subsec:feature_importance}.

\subsection{Economic Performance}
In meme coin copy trading, identifying a profitable wallet is not sufficient to guarantee copier profitability, because execution occurs on a bonding curve where price impact depends on trade ordering. Even when a copier immediately replicates the same trades as a smart money wallet, the copier necessarily trades at a worse point on the curve. This creates a mechanical performance gap that we refer to as the imitation penalty.

To formalize this effect, we derive the returns of smart money and its one-to-one immediate copiers under the pump.fun bonding-curve execution model introduced in \autoref{apx:bonding_curve_mechanism}. The following lemma characterizes their respective returns. \begin{lemma}[Net Returns of Smart Money and Copier]
\label{lemma:return}
Consider a smart-money wallet that executes a sequence of $T$ trades indexed by
$t = 1,\dots,T$, where trade sizes and timing are fixed and not affected by copier behavior. Each trade is characterized by a signed token quantity
$q_t \in \mathbb{R}$, where $q_t > 0$ denotes a buy and $q_t < 0$ denotes a sell.
Let $Y^{S}_{t}$ denote the effective token reserve immediately before the
smart money’s trade $t$. Feasibility requires $Y^{S}_{t} > |q_t|$. The net returns of smart money, $R^{(S)}$, and the corresponding one-to-one immediate copier, $R^{(C)}$, are
{\small$$
R^{(S)} = \frac{-\sum_{t:q_{t}<0} \frac{kq_{t}}{Y^{S}_{t}(Y^{S}_{t}-q_{t})}}{\sum_{t:q_{t}>0} \frac{kq_{t}}{Y^{S}_{t}(Y^{S}_{t}-q_{t})}} - 1,
\,
R^{(C)} = \frac{-\sum_{t:q_{t}<0} \frac{kq_{t}}{(Y^{S}_{t} - q_{t})(Y^{S}_{t}-2q_{t})}}{\sum_{t:q_{t}>0} \frac{kq_{t}}{(Y^{S}_{t} - q_{t})(Y^{S}_{t}-2q_{t})}} - 1.
$$
}
\end{lemma}
\begin{theorem}[Imitation Penalty]
\label{theorem:imitation_penalty}
Under the above execution model, the copier strictly overpays on every replicated buy. Define the imitation penalty as {\small$\epsilon = R^{(S)} - R^{(C)}$}, then, {\small$\epsilon>0$}.
\end{theorem}Proofs of \autoref{lemma:return} and \autoref{theorem:imitation_penalty} are provided in \autoref{apx:leader_copier_return}.

\begin{figure}[tb]
    \centering
    \includegraphics[width=\linewidth]{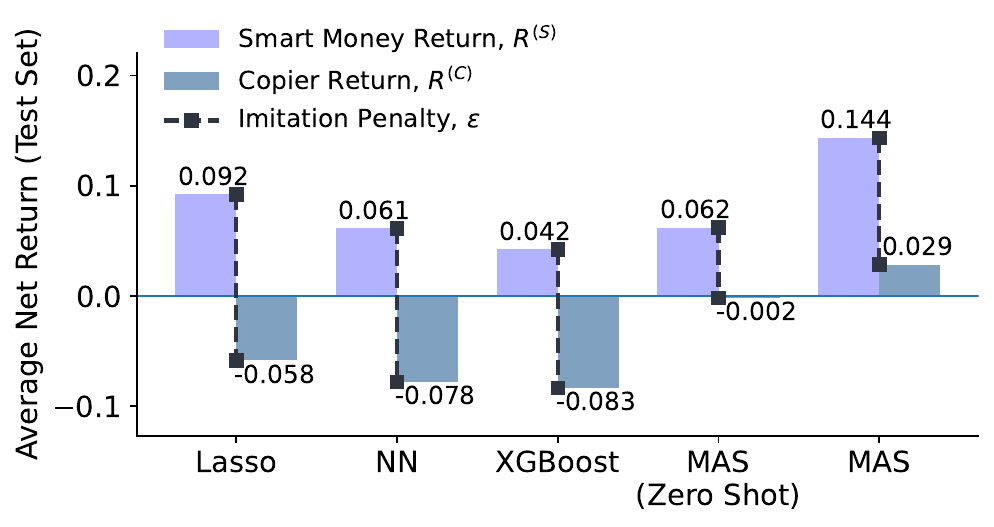}
    \caption{Test set average net return of smart money and copier with market frictions.}
    \label{fig:avg_return_test_bar}
\end{figure}

Based on \autoref{lemma:return} and \autoref{theorem:imitation_penalty}, \autoref{fig:avg_return_test_bar} reports test-set average smart money wallet net returns, corresponding copier net return, and imitation penalty under market frictions. Although smart money returns are positive across all models, copier returns are consistently lower, indicating a persistent imitation penalty arising from price impact. For statistic-driven models, positive smart money returns are insufficient to yield positive copier profitability, with copier returns remaining negative despite accurate wallet identification. In contrast, the \ac{mas}-based model achieves the highest smart money returns and is the only approach that delivers a positive average copier return, demonstrating improved economic effectiveness under realistic market conditions.

\subsection{Ablation Study}

We conduct an ablation study to assess the contribution of each agent in the proposed \ac{mas}-based framework. \autoref{tab:ablation} reports \ac{auc} and gross return (net return $+1$) for both smart money wallets and copiers when removing each agent. Removing the \emph{wallet agent} leads to the largest degradation across all metrics, with an \ac{auc} drop of 11.3\% and pronounced declines in both smart money (6.0\%) and copier returns (9.4\%), underscoring the central role of wallet-level signals in identifying profitable trading behavior. Excluding the \emph{coin agent} also results in substantial performance losses (\ac{auc} down 9.8\%), confirming that token-level characteristics and embedded manipulative behaviors materially affect copy-trading outcomes. In contrast, removing the \emph{timing agent} yields a smaller but non-negligible decline, suggesting that entry timing provides complementary information but is less dominant than wallet- and coin-level assessments.
\begin{table}[tb]
    \centering
    \small
    \caption{Ablation study of the proposed \ac{mas}.}
    \begin{tabularx}{\linewidth}{lccc}
\toprule
\multicolumn{1}{c}{\multirow{2}{*}{\textbf{Ablation}}} &
\multicolumn{1}{c}{\multirow{2}{*}{\textbf{AUC}}} &
\multicolumn{2}{c}{\textbf{Gross Return (Net Return$+1$)}} \\
\cmidrule(lr){3-4}
& & \textbf{Smart Money} & \textbf{Copier} \\
\midrule

w/o Wallet Agent
& $0.6382$\dec{11.3\%}
& $1.0753$\dec{6.0\%}
& $0.9327$\dec{9.4\%} \\

w/o Coin Agent
& $0.6489$\dec{9.8\%}
& $1.0837$\dec{5.2\%}
& $0.9415$\dec{8.5\%} \\

w/o Timing Agent
& $0.6943$\dec{3.5\%}
& $1.1048$\dec{3.4\%}
& $1.0124$\dec{1.6\%} \\

\bottomrule
\end{tabularx}

    \label{tab:ablation}
\end{table}

\section{Conclusion}


In this paper, we propose a \ac{llm}-powered multi-agent framework that explicitly reasons over multi-modal on-chain and off-chain signals. Empirical evaluation on over 6,000 meme coin projects shows that our approach outperforms zero-shot and most statistic-driven baselines in prediction accuracy and all baselines in profitability. Overall, our results demonstrate the effectiveness of agent-based defenses in adversarial meme coin markets and establish that trader profitability is predictable, providing a practical foundation for robust copy trading.
\begin{acks}
The authors gratefully acknowledge Ievgen Gerasymchuk, Wenzhi Ding, and Stefanos Chaliasos for their insightful comments. 

This research is funded by the National Research Foundation, Singapore, and DSO National Laboratories under the AI Singapore Programme (AISG Award No: AISG2-GC-2023-008),
the National Research Foundation, Singapore, and the Cyber Security Agency under its National Cybersecurity R\&D Programme (NCRP25-P04-TAICeN),
the National Research Foundation, Prime Minister’s Office, Singapore under Campus for Research Excellence and Technological Enterprise (CREATE) programme, Engineering and Physical Sciences Research Council (EPSRC) under UKFin+ Programme, and Ripple under University Blockchain Research Initiative (UBRI)~\cite{Feng2022UniversityResearch}.
\end{acks}
\bibliographystyle{ACM-Reference-Format}
\bibliography{references_multi_agent,references_meme}


\begin{thebibliography}{46}


\ifx \showCODEN    \undefined \def \showCODEN     #1{\unskip}     \fi
\ifx \showISBNx    \undefined \def \showISBNx     #1{\unskip}     \fi
\ifx \showISBNxiii \undefined \def \showISBNxiii  #1{\unskip}     \fi
\ifx \showISSN     \undefined \def \showISSN      #1{\unskip}     \fi
\ifx \showLCCN     \undefined \def \showLCCN      #1{\unskip}     \fi
\ifx \shownote     \undefined \def \shownote      #1{#1}          \fi
\ifx \showarticletitle \undefined \def \showarticletitle #1{#1}   \fi
\ifx \showURL      \undefined \def \showURL       {\relax}        \fi
\providecommand\bibfield[2]{#2}
\providecommand\bibinfo[2]{#2}
\providecommand\natexlab[1]{#1}
\providecommand\showeprint[2][]{arXiv:#2}

\bibitem[Aloosh and Li(2024)]%
        {Aloosh2024DirectTrading}
\bibfield{author}{\bibinfo{person}{Arash Aloosh} {and} \bibinfo{person}{Jiasun Li}.} \bibinfo{year}{2024}\natexlab{}.
\newblock \showarticletitle{{Direct Evidence of Bitcoin Wash Trading}}.
\newblock \bibinfo{journal}{\emph{Management Science}} \bibinfo{volume}{70}, \bibinfo{number}{12} (\bibinfo{date}{3} \bibinfo{year}{2024}), \bibinfo{pages}{8875--8921}.
\newblock
\showISSN{15265501}


\bibitem[Apesteguia et~al\mbox{.}(2020)]%
        {Apesteguia2020CopyTrading}
\bibfield{author}{\bibinfo{person}{Jose Apesteguia}, \bibinfo{person}{Jörg Oechssler}, {and} \bibinfo{person}{Simon Weidenholzer}.} \bibinfo{year}{2020}\natexlab{}.
\newblock \showarticletitle{{Copy Trading}}.
\newblock \bibinfo{journal}{\emph{Management Science}} \bibinfo{volume}{66}, \bibinfo{number}{12} (\bibinfo{date}{7} \bibinfo{year}{2020}), \bibinfo{pages}{5608--5622}.
\newblock
\showISSN{15265501}


\bibitem[Cong et~al\mbox{.}(2023)]%
        {Cong2023CryptoTrading}
\bibfield{author}{\bibinfo{person}{Lin~William Cong}, \bibinfo{person}{Xi Li}, \bibinfo{person}{Ke Tang}, {and} \bibinfo{person}{Yang Yang}.} \bibinfo{year}{2023}\natexlab{}.
\newblock \showarticletitle{{Crypto Wash Trading}}.
\newblock \bibinfo{journal}{\emph{Management Science}} \bibinfo{volume}{69}, \bibinfo{number}{11} (\bibinfo{date}{9} \bibinfo{year}{2023}), \bibinfo{pages}{6427--6454}.
\newblock
\showISSN{15265501}


\bibitem[Cruz et~al\mbox{.}(2025)]%
        {Cruz2025AMM-basedLedger}
\bibfield{author}{\bibinfo{person}{Walter~Hernandez Cruz}, \bibinfo{person}{Firas Dahi}, \bibinfo{person}{Yebo Feng}, \bibinfo{person}{Jiahua Xu}, \bibinfo{person}{Aanchal Malhotra}, {and} \bibinfo{person}{Paolo Tasca}.} \bibinfo{year}{2025}\natexlab{}.
\newblock \showarticletitle{{AMM-based DEX on the XRP Ledger}}. In \bibinfo{booktitle}{\emph{IEEE International Conference on Blockchain and Cryptocurrency}}.
\newblock
\showISBNx{9798331541354}


\bibitem[Dhawan and Putniņ{\v{s}}(2023)]%
        {Dhawan2023AMarkets}
\bibfield{author}{\bibinfo{person}{Anirudh Dhawan} {and} \bibinfo{person}{Tālis~J. Putniņ{\v{s}}}.} \bibinfo{year}{2023}\natexlab{}.
\newblock \showarticletitle{{A New Wolf in Town? Pump-and-Dump Manipulation in Cryptocurrency Markets}}.
\newblock \bibinfo{journal}{\emph{Review of Finance}} \bibinfo{volume}{27}, \bibinfo{number}{3} (\bibinfo{date}{5} \bibinfo{year}{2023}), \bibinfo{pages}{935--975}.
\newblock
\showISSN{1572-3097}


\bibitem[Ding et~al\mbox{.}(2025)]%
        {Ding2025DecomposeMarket}
\bibfield{author}{\bibinfo{person}{Wenzhi Ding}, \bibinfo{person}{Chen Lin}, \bibinfo{person}{Yichen Luo}, {and} \bibinfo{person}{Jiahua Xu}.} \bibinfo{year}{2025}\natexlab{}.
\newblock \showarticletitle{{Decompose Market Manipulation Strategies: Evidence from On-chain Meme Coin Market}}.
\newblock \bibinfo{journal}{\emph{SSRN Electronic Journal:5953738}} (\bibinfo{date}{9} \bibinfo{year}{2025}).
\newblock


\bibitem[Erinle et~al\mbox{.}(2025)]%
        {Erinle2025SoK:Wallets}
\bibfield{author}{\bibinfo{person}{Yimika Erinle}, \bibinfo{person}{Yathin Kethepalli}, \bibinfo{person}{Yebo Feng}, {and} \bibinfo{person}{Jiahua Xu}.} \bibinfo{year}{2025}\natexlab{}.
\newblock \showarticletitle{{SoK: Design, vulnerabilities, and security measures of cryptocurrency wallets}}.
\newblock \bibinfo{journal}{\emph{Computer Networks}}  \bibinfo{volume}{273} (\bibinfo{date}{12} \bibinfo{year}{2025}), \bibinfo{pages}{111691}.
\newblock
\showISSN{1389-1286}


\bibitem[Feng et~al\mbox{.}(2022)]%
        {Feng2022UniversityResearch}
\bibfield{author}{\bibinfo{person}{Yebo Feng}, \bibinfo{person}{Jiahua Xu}, {and} \bibinfo{person}{Lauren Weymouth}.} \bibinfo{year}{2022}\natexlab{}.
\newblock \showarticletitle{{University Blockchain Research Initiative (UBRI): Boosting blockchain education and research}}.
\newblock \bibinfo{journal}{\emph{IEEE Potentials}} \bibinfo{volume}{41}, \bibinfo{number}{6} (\bibinfo{year}{2022}), \bibinfo{pages}{19--25}.
\newblock
\showISSN{15581772}


\bibitem[Gao et~al\mbox{.}(2025)]%
        {Gao2025AiRacleX:Generation}
\bibfield{author}{\bibinfo{person}{Bo Gao}, \bibinfo{person}{Yuan Wang}, \bibinfo{person}{Qingsong Wei}, \bibinfo{person}{Yong Liu}, \bibinfo{person}{Rick Siow}, \bibinfo{person}{Mong Goh}, \bibinfo{person}{David Lo}, {and} \bibinfo{person}{Bo Gao}.} \bibinfo{year}{2025}\natexlab{}.
\newblock \showarticletitle{{AiRacleX: Automated Detection of Price Oracle Manipulations via LLM-Driven Knowledge Mining and Prompt Generation}}.
\newblock \bibinfo{journal}{\emph{arXiv:2502.06348}} (\bibinfo{date}{2} \bibinfo{year}{2025}).
\newblock
\showISSN{19391374}


\bibitem[Geng et~al\mbox{.}(2024)]%
        {Geng2024AModels}
\bibfield{author}{\bibinfo{person}{Jiahui Geng}, \bibinfo{person}{Fengyu Cai}, \bibinfo{person}{Yuxia Wang}, \bibinfo{person}{Heinz Koeppl}, \bibinfo{person}{Preslav Nakov}, {and} \bibinfo{person}{Iryna Gurevych}.} \bibinfo{year}{2024}\natexlab{}.
\newblock \showarticletitle{{A Survey of Confidence Estimation and Calibration in Large Language Models}}. In \bibinfo{booktitle}{\emph{Proceedings of the 2024 Conference of the North American Chapter of the Association for Computational Linguistics: Human Language Technologies}}, Vol.~\bibinfo{volume}{1}. \bibinfo{pages}{6577--6595}.
\newblock
\showISBNx{9798891761148}


\bibitem[Kapoor et~al\mbox{.}(2024)]%
        {Kapoor2024LargeKnow}
\bibfield{author}{\bibinfo{person}{Sanyam Kapoor}, \bibinfo{person}{Nate Gruver}, \bibinfo{person}{Manley Roberts}, \bibinfo{person}{Katherine Collins}, \bibinfo{person}{Arka Pal}, \bibinfo{person}{Umang Bhatt}, \bibinfo{person}{Adrian Weller}, \bibinfo{person}{Samuel Dooley}, \bibinfo{person}{Micah Goldblum}, {and} \bibinfo{person}{Andrew~Gordon Wilson}.} \bibinfo{year}{2024}\natexlab{}.
\newblock \showarticletitle{{Large Language Models Must Be Taught to Know What They Don’t Know}}. In \bibinfo{booktitle}{\emph{Proceedings of the 38th International Conference on Neural Information Processing Systems}}, Vol.~\bibinfo{volume}{37}. \bibinfo{pages}{85932--85972}.
\newblock
\showISSN{10495258}


\bibitem[Kawai et~al\mbox{.}(2024)]%
        {Kawai2024StrangerPlatforms}
\bibfield{author}{\bibinfo{person}{Daisuke Kawai}, \bibinfo{person}{Kyle Soska}, \bibinfo{person}{Bryan Routledge}, \bibinfo{person}{Ariel Zetlin-Jones}, {and} \bibinfo{person}{Nicolas Christin}.} \bibinfo{year}{2024}\natexlab{}.
\newblock \showarticletitle{{Stranger Danger? Investor Behavior and Incentives on Cryptocurrency Copy-Trading Platforms}}. In \bibinfo{booktitle}{\emph{Proceedings of the CHI Conference on Human Factors in Computing Systems}}. \bibinfo{address}{New York, NY, USA}, \bibinfo{pages}{1--20}.
\newblock
\showISBNx{9798400703300}


\bibitem[Khanmohammadi et~al\mbox{.}(2025)]%
        {Khanmohammadi2025CalibratingStability}
\bibfield{author}{\bibinfo{person}{Reza Khanmohammadi}, \bibinfo{person}{Erfan Miahi}, \bibinfo{person}{Mehrsa Mardikoraem}, \bibinfo{person}{Simerjot Kaur}, \bibinfo{person}{Ivan Brugere}, \bibinfo{person}{Charese~H Smiley}, \bibinfo{person}{Kundan Thind}, \bibinfo{person}{Mohammad~M Ghassemi}, \bibinfo{person}{Jpmorgan~Ai Research}, {and} \bibinfo{person}{Henry Ford~Health}.} \bibinfo{year}{2025}\natexlab{}.
\newblock \showarticletitle{{Calibrating LLM Confidence by Probing Perturbed Representation Stability}}. In \bibinfo{booktitle}{\emph{Proceedings of the 2025 Conference on Empirical Methods in Natural Language Processing}}. \bibinfo{pages}{10448--10514}.
\newblock


\bibitem[Li et~al\mbox{.}(2024)]%
        {Li2024CryptoTrade:Trading}
\bibfield{author}{\bibinfo{person}{Yuan Li}, \bibinfo{person}{Bingqiao Luo}, \bibinfo{person}{Qian Wang}, \bibinfo{person}{Nuo Chen}, \bibinfo{person}{Xu Liu}, {and} \bibinfo{person}{Bingsheng He}.} \bibinfo{year}{2024}\natexlab{}.
\newblock \showarticletitle{{CryptoTrade: A Reflective LLM-based Agent to Guide Zero-shot Cryptocurrency Trading}}. In \bibinfo{booktitle}{\emph{Empirical Methods in Natural Language Processing}}. \bibinfo{pages}{1094--1106}.
\newblock
\showISBNx{9798891761643}


\bibitem[Li et~al\mbox{.}(2023)]%
        {Li2023LargeSurvey}
\bibfield{author}{\bibinfo{person}{Yinheng Li}, \bibinfo{person}{Shaofei Wang}, \bibinfo{person}{Han Ding}, {and} \bibinfo{person}{Hang Chen}.} \bibinfo{year}{2023}\natexlab{}.
\newblock \showarticletitle{{Large Language Models in Finance: A Survey}}.
\newblock \bibinfo{journal}{\emph{arXiv:2311.10723}} (\bibinfo{year}{2023}).
\newblock
\showISBNx{9798400702402}


\bibitem[Li et~al\mbox{.}(2025)]%
        {Li2025TrustMarkets}
\bibfield{author}{\bibinfo{person}{Yueyao Li}, \bibinfo{person}{Nanjun Yao}, \bibinfo{person}{Yuhui Huo}, {and} \bibinfo{person}{Wei Cai}.} \bibinfo{year}{2025}\natexlab{}.
\newblock \showarticletitle{{Trust Dynamics and Bot-Driven Responses: An Approach to Rug Pulls in Solana Meme Coin Markets}}. In \bibinfo{booktitle}{\emph{Proceedings of the 17th ACM Web Science Conference}}, Vol.~\bibinfo{volume}{25}. \bibinfo{pages}{106--116}.
\newblock
\showISBNx{9798400714832}


\bibitem[Liu et~al\mbox{.}(2023)]%
        {Liu2023ChainFeedback}
\bibfield{author}{\bibinfo{person}{Hao Liu}, \bibinfo{person}{Carmelo Sferrazza}, {and} \bibinfo{person}{Pieter Abbeel}.} \bibinfo{year}{2023}\natexlab{}.
\newblock \showarticletitle{{Chain of Hindsight Aligns Language Models with Feedback}}. In \bibinfo{booktitle}{\emph{12th International Conference on Learning Representations}}.
\newblock


\bibitem[Long et~al\mbox{.}(2025a)]%
        {Long2025CoinCLIP:Memecoins}
\bibfield{author}{\bibinfo{person}{Houwan Long}, \bibinfo{person}{Hongyang Li}, {and} \bibinfo{person}{Wei Cai}.} \bibinfo{year}{2025}\natexlab{a}.
\newblock \showarticletitle{{CoinCLIP: A Multimodal Framework for Assessing Viability in Web3 Memecoins}}. In \bibinfo{booktitle}{\emph{Proceedings of the 34th ACM International Conference on Information and Knowledge Management}}. \bibinfo{pages}{4986--4990}.
\newblock
\showISBNx{9798400720406}


\bibitem[Long et~al\mbox{.}(2025b)]%
        {Long2025BridgingEcosystem}
\bibfield{author}{\bibinfo{person}{Hou~Wan Long}, \bibinfo{person}{Nga~Man Wong}, {and} \bibinfo{person}{Wei Cai}.} \bibinfo{year}{2025}\natexlab{b}.
\newblock \showarticletitle{{Bridging Culture and Finance: A Multimodal Analysis of Memecoins in the Web3 Ecosystem}}. In \bibinfo{booktitle}{\emph{Companion Proceedings of the ACM Web Conference 2025}}. \bibinfo{pages}{1158--1161}.
\newblock
\showISBNx{9798400713316}


\bibitem[Lundberg and Lee(2017)]%
        {Lundberg2017APredictions}
\bibfield{author}{\bibinfo{person}{Scott Lundberg} {and} \bibinfo{person}{Su-In Lee}.} \bibinfo{year}{2017}\natexlab{}.
\newblock \showarticletitle{{A Unified Approach to Interpreting Model Predictions}}. In \bibinfo{booktitle}{\emph{Proceedings of the 31st International Conference on Neural Information Processing Systems}}, Vol.~\bibinfo{volume}{2017-December}. \bibinfo{pages}{4768--4777}.
\newblock


\bibitem[Luo et~al\mbox{.}(2026)]%
        {Luo2026PiercingReappraised}
\bibfield{author}{\bibinfo{person}{Yichen Luo}, \bibinfo{person}{Yebo Feng}, \bibinfo{person}{Jiahua Xu}, {and} \bibinfo{person}{Paolo Tasca}.} \bibinfo{year}{2026}\natexlab{}.
\newblock \showarticletitle{{Piercing the Veil of TVL: DeFi Reappraised}}. In \bibinfo{booktitle}{\emph{Financial Cryptography and Data Security}}. \bibinfo{pages}{3--19}.
\newblock
\showISBNx{978-3-032-07035-7}
\showISSN{1611-3349}


\bibitem[Luo et~al\mbox{.}(2025)]%
        {Luo2025LLM-PoweredManagement}
\bibfield{author}{\bibinfo{person}{Yichen Luo}, \bibinfo{person}{Yebo Feng}, \bibinfo{person}{Jiahua Xu}, \bibinfo{person}{Paolo Tasca}, {and} \bibinfo{person}{Yang Liu}.} \bibinfo{year}{2025}\natexlab{}.
\newblock \showarticletitle{{LLM-Powered Multi-Agent System for Automated Crypto Portfolio Management}}.
\newblock \bibinfo{journal}{\emph{arXiv:2501.00826}} (\bibinfo{date}{1} \bibinfo{year}{2025}).
\newblock


\bibitem[Mao et~al\mbox{.}(2025)]%
        {Mao2025KnowMining}
\bibfield{author}{\bibinfo{person}{Qian'ang Mao}, \bibinfo{person}{Yuxuan Zhang}, \bibinfo{person}{Jiaman Chen}, \bibinfo{person}{Wenjun Zhou}, {and} \bibinfo{person}{Jiaqi Yan}.} \bibinfo{year}{2025}\natexlab{}.
\newblock \showarticletitle{{Know Your Intent: An Autonomous Multi-Perspective LLM Agent Framework for DeFi User Transaction Intent Mining}}.
\newblock \bibinfo{journal}{\emph{arXiv:2511.15456}}  \bibinfo{volume}{1} (\bibinfo{year}{2025}).
\newblock


\bibitem[Mazorra et~al\mbox{.}(2022)]%
        {Mazorra2022DoDetection}
\bibfield{author}{\bibinfo{person}{Bruno Mazorra}, \bibinfo{person}{Victor Adan}, {and} \bibinfo{person}{Vanesa Daza}.} \bibinfo{year}{2022}\natexlab{}.
\newblock \showarticletitle{{Do Not Rug on Me: Leveraging Machine Learning Techniques for Automated Scam Detection}}.
\newblock \bibinfo{journal}{\emph{Mathematics}} \bibinfo{volume}{10}, \bibinfo{number}{6} (\bibinfo{date}{3} \bibinfo{year}{2022}), \bibinfo{pages}{949}.
\newblock
\showISSN{2227-7390}


\bibitem[Mongardini and Mei(2026)]%
        {Mongardini2026AEcosystem}
\bibfield{author}{\bibinfo{person}{Alberto~Maria Mongardini} {and} \bibinfo{person}{Alessandro Mei}.} \bibinfo{year}{2026}\natexlab{}.
\newblock \showarticletitle{{A Midsummer Meme's Dream: Investigating Market Manipulations in the Meme Coin Ecosystem}}.
\newblock \bibinfo{journal}{\emph{arXiv:2507.01963}} (\bibinfo{date}{1} \bibinfo{year}{2026}).
\newblock


\bibitem[Nguyen et~al\mbox{.}(2023)]%
        {Nguyen2023Rug-pullEngineering}
\bibfield{author}{\bibinfo{person}{Minh~Hoang Nguyen}, \bibinfo{person}{Phuong~Duy Huynh}, \bibinfo{person}{Son~Hoang Dau}, {and} \bibinfo{person}{Xiaodong Li}.} \bibinfo{year}{2023}\natexlab{}.
\newblock \showarticletitle{{Rug-pull malicious token detection on blockchain using supervised learning with feature engineering}}. In \bibinfo{booktitle}{\emph{Australasian Computer Science Week}}. \bibinfo{pages}{72--81}.
\newblock
\showISBNx{9798400700057}


\bibitem[Nitarach et~al\mbox{.}(2025)]%
        {Nitarach2025FinCoT:Reasoning}
\bibfield{author}{\bibinfo{person}{Natapong Nitarach}, \bibinfo{person}{Warit Sirichotedumrong}, \bibinfo{person}{Panop Pitchayarthorn}, \bibinfo{person}{Pittawat Taveekitworachai}, \bibinfo{person}{Potsawee Manakul}, {and} \bibinfo{person}{Kunat Pipatanakul}.} \bibinfo{year}{2025}\natexlab{}.
\newblock \showarticletitle{{FinCoT: Grounding Chain-of-Thought in Expert Financial Reasoning}}.
\newblock \bibinfo{journal}{\emph{arXiv:2506.16123}} (\bibinfo{date}{6} \bibinfo{year}{2025}).
\newblock


\bibitem[Niu et~al\mbox{.}(2024)]%
        {Niu2024UnveilingMarkets}
\bibfield{author}{\bibinfo{person}{Yuanzheng Niu}, \bibinfo{person}{Xiaoqi Li}, \bibinfo{person}{Hongli Peng}, {and} \bibinfo{person}{Wenkai Li}.} \bibinfo{year}{2024}\natexlab{}.
\newblock \showarticletitle{{Unveiling Wash Trading in Popular NFT Markets}}. In \bibinfo{booktitle}{\emph{Companion Proceedings of the ACM Web Conference}}. \bibinfo{pages}{730--733}.
\newblock
\showISBNx{9798400701726}


\bibitem[Perez et~al\mbox{.}(2020)]%
        {Perez2020RevisitingBlockchains}
\bibfield{author}{\bibinfo{person}{Daniel Perez}, \bibinfo{person}{Jiahua Xu}, {and} \bibinfo{person}{Benjamin Livshits}.} \bibinfo{year}{2020}\natexlab{}.
\newblock \showarticletitle{{Revisiting Transactional Statistics of High-scalability Blockchains}}. In \bibinfo{booktitle}{\emph{Proceedings of the ACM SIGCOMM Internet Measurement Conference}}. \bibinfo{pages}{535--550}.
\newblock
\showISBNx{9781450381383}


\bibitem[Shen et~al\mbox{.}(2025)]%
        {Shen2025Irrationality-AwareTrading}
\bibfield{author}{\bibinfo{person}{Zhe Shen}, \bibinfo{person}{Wei Jiang}, {and} \bibinfo{person}{Zhiqiang~(Eric) Zheng}.} \bibinfo{year}{2025}\natexlab{}.
\newblock \showarticletitle{{Irrationality-Aware Human Machine Collaboration: Mitigating Alterfactual Irrationality in Copy Trading}}.
\newblock \bibinfo{journal}{\emph{Information System Research}} (\bibinfo{date}{4} \bibinfo{year}{2025}).
\newblock
\showISSN{1047-7047}


\bibitem[Sun et~al\mbox{.}(2025)]%
        {Sun2025SoK:Assessment}
\bibfield{author}{\bibinfo{person}{Dianxiang Sun}, \bibinfo{person}{Wei Ma}, \bibinfo{person}{Liming Nie}, {and} \bibinfo{person}{Yang Liu}.} \bibinfo{year}{2025}\natexlab{}.
\newblock \showarticletitle{{SoK: A Taxonomic Analysis of DeFi Rug Pulls: Types, Dataset, and Tool Assessment}}.
\newblock \bibinfo{journal}{\emph{Proceedings of the ACM on Software Engineering}} \bibinfo{volume}{2}, \bibinfo{number}{ISSTA} (\bibinfo{date}{6} \bibinfo{year}{2025}), \bibinfo{pages}{550--572}.
\newblock
\showISSN{2994-970X}


\bibitem[Wang et~al\mbox{.}(2024)]%
        {Wang2024ExploringReasoning}
\bibfield{author}{\bibinfo{person}{Qian Wang}, \bibinfo{person}{Yuchen Gao}, \bibinfo{person}{Zhenheng Tang}, \bibinfo{person}{Bingqiao Luo}, \bibinfo{person}{Nuo Chen}, {and} \bibinfo{person}{Bingsheng He}.} \bibinfo{year}{2024}\natexlab{}.
\newblock \showarticletitle{{Exploring LLM Cryptocurrency Trading Through Fact-Subjectivity Aware Reasoning}}. In \bibinfo{booktitle}{\emph{ICLR 2025 Advances in Financial AI Workshop}}.
\newblock


\bibitem[Wei et~al\mbox{.}(2022)]%
        {Wei2022Chain-of-ThoughtModels}
\bibfield{author}{\bibinfo{person}{Jason Wei}, \bibinfo{person}{Xuezhi Wang}, \bibinfo{person}{Dale Schuurmans}, \bibinfo{person}{Maarten Bosma}, \bibinfo{person}{brian ichter}, \bibinfo{person}{Fei Xia}, \bibinfo{person}{Ed Chi}, \bibinfo{person}{Quoc~V Le}, {and} \bibinfo{person}{Denny Zhou}.} \bibinfo{year}{2022}\natexlab{}.
\newblock \showarticletitle{{Chain-of-Thought Prompting Elicits Reasoning in Large Language Models}}. In \bibinfo{booktitle}{\emph{Proceedings of the 36th International Conference on Neural Information Processing Systems}}, Vol.~\bibinfo{volume}{35}. \bibinfo{pages}{24824--24837}.
\newblock


\bibitem[Wei et~al\mbox{.}(2025)]%
        {Wei2025AdvancedSystems}
\bibfield{author}{\bibinfo{person}{Zhiyuan Wei}, \bibinfo{person}{Jing Sun}, \bibinfo{person}{Yuqiang Sun}, \bibinfo{person}{Ye Liu}, \bibinfo{person}{Daoyuan Wu}, \bibinfo{person}{Zijian Zhang}, \bibinfo{person}{Xianhao Zhang}, \bibinfo{person}{Meng Li}, \bibinfo{person}{Yang Liu}, \bibinfo{person}{Chunmiao Li}, \bibinfo{person}{Mingchao Wan}, \bibinfo{person}{Jin Dong}, {and} \bibinfo{person}{Liehuang Zhu}.} \bibinfo{year}{2025}\natexlab{}.
\newblock \showarticletitle{{Advanced Smart Contract Vulnerability Detection via LLM-Powered Multi-Agent Systems}}.
\newblock \bibinfo{journal}{\emph{IEEE Transactions on Software Engineering}} \bibinfo{volume}{51}, \bibinfo{number}{10} (\bibinfo{date}{10} \bibinfo{year}{2025}), \bibinfo{pages}{2830--2846}.
\newblock
\showISSN{19393520}


\bibitem[Wu et~al\mbox{.}(2025)]%
        {Wu2025ProfitLearning}
\bibfield{author}{\bibinfo{person}{Cong Wu}, \bibinfo{person}{Jing Chen}, \bibinfo{person}{Jiahong Li}, \bibinfo{person}{Jiahua Xu}, \bibinfo{person}{Ju Jia}, \bibinfo{person}{Yutao Hu}, \bibinfo{person}{Yebo Feng}, \bibinfo{person}{Yang Liu}, {and} \bibinfo{person}{Yang Xiang}.} \bibinfo{year}{2025}\natexlab{}.
\newblock \showarticletitle{{Profit or Deceit? Mitigating Pump and Dump in DeFi via Graph and Contrastive Learning}}.
\newblock \bibinfo{journal}{\emph{IEEE Transactions on Information Forensics and Security}}  \bibinfo{volume}{20} (\bibinfo{year}{2025}), \bibinfo{pages}{8994--9008}.
\newblock
\showISSN{15566021}


\bibitem[Xiang et~al\mbox{.}(2025)]%
        {Xiang2025MeasuringFragility}
\bibfield{author}{\bibinfo{person}{Yuexin Xiang}, \bibinfo{person}{Qishuang Fu}, \bibinfo{person}{Yuquan Li}, \bibinfo{person}{Qin Wang}, \bibinfo{person}{Tsz~Hon Yuen}, {and} \bibinfo{person}{Jiangshan Yu}.} \bibinfo{year}{2025}\natexlab{}.
\newblock \showarticletitle{{Measuring Memecoin Fragility}}.
\newblock \bibinfo{journal}{\emph{arXiv:2512.00377}} (\bibinfo{date}{11} \bibinfo{year}{2025}).
\newblock


\bibitem[Xie et~al\mbox{.}(2025)]%
        {Xie2025FinChain:Reasoning}
\bibfield{author}{\bibinfo{person}{Zhuohan Xie}, \bibinfo{person}{Daniil Orel}, \bibinfo{person}{Rushil Thareja}, \bibinfo{person}{Dhruv Sahnan}, \bibinfo{person}{Hachem Madmoun}, \bibinfo{person}{Fan Zhang}, \bibinfo{person}{Debopriyo Banerjee}, \bibinfo{person}{Georgi Georgiev}, \bibinfo{person}{Xueqing Peng}, \bibinfo{person}{Lingfei Qian}, \bibinfo{person}{Jimin Huang}, \bibinfo{person}{Jinyan Su}, \bibinfo{person}{Aaryamonvikram Singh}, \bibinfo{person}{Rui Xing}, \bibinfo{person}{Rania Elbadry}, \bibinfo{person}{Chen Xu}, \bibinfo{person}{Haonan Li}, \bibinfo{person}{Fajri Koto}, \bibinfo{person}{Ivan Koychev}, \bibinfo{person}{Tanmoy Chakraborty}, \bibinfo{person}{Yuxia Wang}, \bibinfo{person}{Salem Lahlou}, \bibinfo{person}{Veselin Stoyanov}, \bibinfo{person}{Sophia Ananiadou}, \bibinfo{person}{Preslav Nakov}, {and} \bibinfo{person}{St Kliment~Ohridski}.} \bibinfo{year}{2025}\natexlab{}.
\newblock \showarticletitle{{FinChain: A Symbolic Benchmark for Verifiable Chain-of-Thought Financial Reasoning}}.
\newblock \bibinfo{journal}{\emph{arXiv:2506.02515}} (\bibinfo{date}{6} \bibinfo{year}{2025}).
\newblock


\bibitem[Xiong et~al\mbox{.}(2025a)]%
        {Xiong2025EnhancingAggregation}
\bibfield{author}{\bibinfo{person}{Siheng Xiong}, \bibinfo{person}{Ali Payani}, {and} \bibinfo{person}{Faramarz Fekri}.} \bibinfo{year}{2025}\natexlab{a}.
\newblock \showarticletitle{{Enhancing Long Chain-of-Thought Reasoning through Multi-Path Plan Aggregation}}.
\newblock \bibinfo{journal}{\emph{arXiv:2510.11620}} (\bibinfo{date}{10} \bibinfo{year}{2025}).
\newblock


\bibitem[Xiong et~al\mbox{.}(2025b)]%
        {Xiong2025DeliberateModel}
\bibfield{author}{\bibinfo{person}{Siheng Xiong}, \bibinfo{person}{Ali Payani}, \bibinfo{person}{Yuan Yang}, {and} \bibinfo{person}{Faramarz Fekri}.} \bibinfo{year}{2025}\natexlab{b}.
\newblock \showarticletitle{{Deliberate Reasoning in Language Models as Structure-Aware Planning with an Accurate World Model}}.
\newblock \bibinfo{journal}{\emph{Proceedings of the Annual Meeting of the Association for Computational Linguistics}}  \bibinfo{volume}{1} (\bibinfo{year}{2025}), \bibinfo{pages}{31900--31931}.
\newblock
\showISBNx{9798891762510}
\showISSN{0736587X}


\bibitem[Xu et~al\mbox{.}(2025)]%
        {Xu2025Auto.gov:DeFi}
\bibfield{author}{\bibinfo{person}{Jiahua Xu}, \bibinfo{person}{Yebo Feng}, \bibinfo{person}{Daniel Perez}, {and} \bibinfo{person}{Benjamin Livshits}.} \bibinfo{year}{2025}\natexlab{}.
\newblock \showarticletitle{{Auto.gov: Learning-Based Governance for Decentralized Finance (DeFi)}}.
\newblock \bibinfo{journal}{\emph{IEEE Transactions on Services Computing}} \bibinfo{volume}{18}, \bibinfo{number}{3} (\bibinfo{year}{2025}), \bibinfo{pages}{1278--1292}.
\newblock
\showISSN{19391374}


\bibitem[Xu and Livshits(2019)]%
        {Xu2019TheScheme}
\bibfield{author}{\bibinfo{person}{Jiahua Xu} {and} \bibinfo{person}{Benjamin Livshits}.} \bibinfo{year}{2019}\natexlab{}.
\newblock \showarticletitle{{The Anatomy of a Cryptocurrency Pump-and-Dump Scheme}}. In \bibinfo{booktitle}{\emph{Proceedings of the 28th USENIX Conference on Security Symposium}}. \bibinfo{pages}{1609--1625}.
\newblock


\bibitem[Xu et~al\mbox{.}(2023)]%
        {Xu2023SoK:Protocols}
\bibfield{author}{\bibinfo{person}{Jiahua Xu}, \bibinfo{person}{Krzysztof Paruch}, \bibinfo{person}{Simon Cousaert}, {and} \bibinfo{person}{Yebo Feng}.} \bibinfo{year}{2023}\natexlab{}.
\newblock \showarticletitle{{SoK: Decentralized Exchanges (DEX) with Automated Market Maker (AMM) Protocols}}.
\newblock \bibinfo{journal}{\emph{Comput. Surveys}} \bibinfo{volume}{55}, \bibinfo{number}{11} (\bibinfo{date}{11} \bibinfo{year}{2023}).
\newblock
\showISSN{15577341}


\bibitem[Yang et~al\mbox{.}(2024)]%
        {Yang2024HarnessingBeyond}
\bibfield{author}{\bibinfo{person}{Jingfeng Yang}, \bibinfo{person}{Hongye Jin}, \bibinfo{person}{Ruixiang Tang}, \bibinfo{person}{Xiaotian Han}, \bibinfo{person}{Qizhang Feng}, \bibinfo{person}{Haoming Jiang}, \bibinfo{person}{Shaochen Zhong}, \bibinfo{person}{Bing Yin}, {and} \bibinfo{person}{Xia Hu}.} \bibinfo{year}{2024}\natexlab{}.
\newblock \showarticletitle{{Harnessing the Power of LLMs in Practice: A Survey on ChatGPT and Beyond}}.
\newblock \bibinfo{journal}{\emph{ACM Transactions on Knowledge Discovery from Data}} \bibinfo{volume}{18}, \bibinfo{number}{6} (\bibinfo{date}{7} \bibinfo{year}{2024}), \bibinfo{pages}{1--32}.
\newblock
\showISSN{1556-4681}


\bibitem[Yin et~al\mbox{.}(2024)]%
        {Yin2024AModels}
\bibfield{author}{\bibinfo{person}{Shukang Yin}, \bibinfo{person}{Chaoyou Fu}, \bibinfo{person}{Sirui Zhao}, \bibinfo{person}{Ke Li}, \bibinfo{person}{Xing Sun}, \bibinfo{person}{Tong Xu}, {and} \bibinfo{person}{Enhong Chen}.} \bibinfo{year}{2024}\natexlab{}.
\newblock \showarticletitle{{A Survey on Multimodal Large Language Models}}.
\newblock \bibinfo{journal}{\emph{National Science Review}} (\bibinfo{date}{11} \bibinfo{year}{2024}).
\newblock
\showISSN{2095-5138}


\bibitem[Zhang et~al\mbox{.}(2024)]%
        {Zhang2024Vision-LanguageSurvey}
\bibfield{author}{\bibinfo{person}{Jingyi Zhang}, \bibinfo{person}{Jiaxing Huang}, \bibinfo{person}{Sheng Jin}, {and} \bibinfo{person}{Shijian Lu}.} \bibinfo{year}{2024}\natexlab{}.
\newblock \showarticletitle{{Vision-Language Models for Vision Tasks: A Survey}}.
\newblock \bibinfo{journal}{\emph{IEEE Transactions on Pattern Analysis and Machine Intelligence}} \bibinfo{volume}{46}, \bibinfo{number}{8} (\bibinfo{year}{2024}), \bibinfo{pages}{5625--5644}.
\newblock
\showISSN{19393539}


\bibitem[Zhong et~al\mbox{.}(2025)]%
        {Zhong2025DetectingReasoning}
\bibfield{author}{\bibinfo{person}{Juantao Zhong}, \bibinfo{person}{Daoyuan Wu}, \bibinfo{person}{Ye Liu}, \bibinfo{person}{Maoyi Xie}, \bibinfo{person}{Yang Liu}, \bibinfo{person}{Yi Li}, {and} \bibinfo{person}{Ning Liu}.} \bibinfo{year}{2025}\natexlab{}.
\newblock \showarticletitle{{Detecting Various DeFi Price Manipulations with LLM Reasoning}}. In \bibinfo{booktitle}{\emph{Proceedings of IEEE/ACM Automated Software Engineering}}.
\newblock


\end{thebibliography}
\appendix
\section{Appendix}
\subsection{Bonding Curve Mechanism}
\label{apx:bonding_curve_mechanism}
Pump.fun adopts a constant-product bonding curve to govern the relationship between deposited SOL and received tokens~\cite{Xu2023SoK:Protocols,Cruz2025AMM-basedLedger}:
\begin{equation}
    \small
    \label{eq:sol_meme}
    y = y' - \frac{k}{x + x'},
\end{equation}where $x$ is the amount of SOL deposited, and $y$ is the corresponding number of meme coins issued. The parameters $x', y' > 0$ denote protocol-fixed virtual reserves.

Define effective reserves $X \equiv x' + x$ and $Y \equiv y' - y$, so that the curve satisfies the invariant $XY=k=x'y'$.

\paragraph{Monotonicity and concavity.}
Taking derivatives with respect to cumulative demand $x$, we obtain
{\small
$$
\frac{d y}{d x} = \frac{k}{(x'+x)^2} > 0,
\qquad
\frac{d^2 y}{d x^2} = -\frac{2k}{(x'+x)^3} < 0,
$$
}
for all $x \ge 0$, since $k>0$ and $x'>0$. Hence, the issuance function $y(x)$ is strictly increasing and strictly concave: additional SOL deposits always yield more tokens, but at a diminishing rate.

The marginal price of the meme coin, denominated in SOL, is the reciprocal slope of the issuance function:
\begin{equation}
    \small
    \label{eq:sol_price}
    p = \frac{dx}{dy} = \frac{(x' + x)^{2}}{k} = \frac{k}{(y' - y)^{2}}.
\end{equation}

Differentiating yields
{\small
$$
\frac{d p}{d x} = \frac{2(x'+x)}{k} > 0,
\qquad
\frac{d^2 p}{d x^2} = \frac{2}{k} > 0,
$$
}
implying that the price is strictly increasing and strictly convex in cumulative demand.

\subsection{Comment Bot Detection Prompt}
\label{apx:comment_bot_detection_prompt}

To complete the comment bot detection task in \autoref{subsubsec:comment_bot}, we design a prompt to classify whether a given comment is human-generated or bot-generated based on its semantic content. The prompt provides the \ac{llm} with a small set of manually curated few-shot examples, including both human-generated comments and bot-generated comments exhibiting contextless, hype-driven, or slogan-like language. Each target comment is then presented independently, and the \ac{llm} outputs a binary classification indicating whether the comment is bot-generated. The prompt template is provided in Prompt~\autoref{pmt:comment_bot}.
\begin{prompt}[label={pmt:comment_bot}]{Comment Bot Detection.}
You are a meme coin comment analyzer. Your task is to classify a given comment as bot-generated or human-generated. Bot-generated comments are often short, context-less, and mass-producible slogans that express hype or hostility. In contrast, human-generated comments tend to be more personalized and nuanced, containing context or opinion with reasoning. Comments that reference other users (e.g., @89009679) are typically human-generated, although not all human comments contain such references. Respond with id and true (if the comment is bot-generated) or false (if it is human-generated). Your response should follow this format:`\{``result": <true/false>\}'
\begin{minipage}[t]{0.60\linewidth}
TO THE MOON!!! READYY

we’ll get there! LFG 

@88857219 show screenshot as proof pls?

Fake web bros, not same ca

\textcolor{blue}{\{\textit{comment}\}}
\end{minipage}
\hfill
\begin{minipage}[t]{0.39\linewidth}
\raggedleft
`\{``result": true\}'

`\{``result": true\}'

`\{``result": false\}'

`\{``result": false\}'
\end{minipage}
\end{prompt}

\subsection{Preprocessing and Hyperparameter Tuning for Statistic-driven Models}
\label{apx:hyperparameters}

This section documents the preprocessing procedures and hyperparameter tuning used for the statistic-driven models. Continuous features are first winsorized at the 2.5\% and 97.5\% percentiles to mitigate the influence of outliers, and subsequently standardized to zero mean and unit variance using statistics computed from the training set. The same preprocessing transformations are then applied to the validation and test sets. All model hyperparameters are tuned using the training and validation sets, and the corresponding hyperparameter search spaces are summarized in \autoref{tab:hyperparams}.
\begin{table}[t]
    \centering
    \footnotesize
    \caption{Hyperparameters of statistic-driven models.}
    \begin{tabular}{p{0.04\columnwidth}p{0.04\columnwidth}p{0.65\columnwidth}}
\toprule
\textbf{Model} & \multicolumn{1}{c}{\textbf{Hyperparams}} & \textbf{Search Space} \\
\midrule
\multicolumn{1}{c}{\multirow{3}{*}{\rotatebox[origin=c]{90}{\textbf{LASSO}}}}
& \multicolumn{1}{|l}{$C$} & $\{10^{1},\dots,10^{4}\}$ \\
& \multicolumn{1}{|l}{Solver} & liblinear \\
& \multicolumn{1}{|l}{Class weight} & balanced \\

\midrule

\multicolumn{1}{c}{\multirow{4}{*}{\rotatebox[origin=c]{90}{\textbf{NN}}}}
& \multicolumn{1}{|l}{Hidden layers} & $\{(32),(32,16),(32,16,8),(32,16,8,4),(32,16,8,4,2)\}$ \\
& \multicolumn{1}{|l}{$\alpha$} & $\{10^{-5},10^{-4},10^{-3}\}$ \\
& \multicolumn{1}{|l}{Learning rate $\eta$}  & $\{10^{-4},5{\times}10^{-4},10^{-3},10^{-2}\}$ \\
& \multicolumn{1}{|l}{Optimizer} & Adam \\

\midrule

\multicolumn{1}{c}{\multirow{7}{*}{\rotatebox[origin=c]{90}{\textbf{XGBoost}}}}
& \multicolumn{1}{|l}{$n_{\text{estimators}}$} & $\{300,600\}$ \\
& \multicolumn{1}{|l}{Max depth} & $\{2,3,4,6\}$ \\
& \multicolumn{1}{|l}{Learning rate} & $\{0.01,0.05,0.1\}$ \\
& \multicolumn{1}{|l}{Subsample} & $\{0.7,0.9,1.0\}$ \\
& \multicolumn{1}{|l}{Column sample} & $\{0.7,0.9,1.0\}$ \\
& \multicolumn{1}{|l}{Min child weight} & $\{1,5,10\}$ \\
& \multicolumn{1}{|l}{$\lambda$ (L2)} & $\{1,5,10\}$ \\
\bottomrule
\end{tabular}

    \label{tab:hyperparams}
\end{table}

\begin{figure}[!h]
    \centering
    \includegraphics[width=\linewidth]{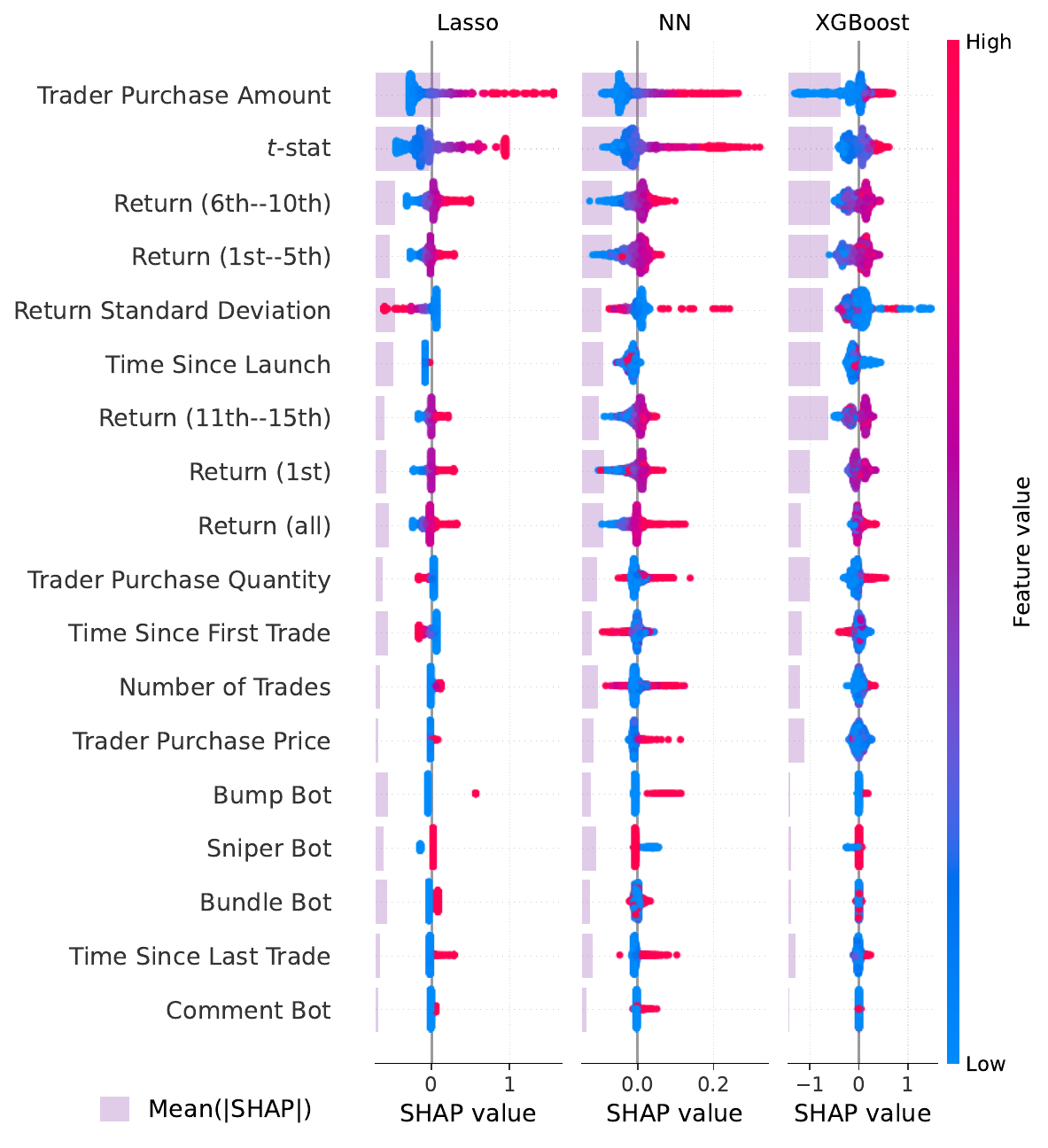}
    \caption{\acs{shap} summary computed on the validation set.}
    \label{tab:feature_importance}
\end{figure}
\subsection{Feature Importance}
\label{subsec:feature_importance}
In this section, we measure feature importance using \ac{shap} on the validation set~\cite{Lundberg2017APredictions}, as summarized in \autoref{tab:feature_importance}. For LASSO, importance is computed using \ac{shap} LinearExplainer values. For XGBoost, we use \ac{shap} TreeExplainer to derive feature attributions. For the \ac{nn}, we compute \ac{shap} values using PermutationExplainer.

\subsection{Few-shot \acp{cot} and Prompts for the \ac{mas}}
\label{apx:json_cot_pmt}

This section describes how we employ few-shot \ac{cot} prompting to guide the wallet, coin, and timing agents within the proposed \ac{mas}.

\subsubsection{\ac{cot}}
\label{apx:cot}
In this section, we present the few-shot \ac{cot} exemplars used to guide agent reasoning, as provided in \ac{cot} \autoref{cot:wallet_agent}, \autoref{cot:coin_agent}, and \autoref{cot:timing_agent}. Each exemplar demonstrates how domain-specific signals should be interpreted, how intermediate reasoning steps should be structured, and how final decisions should be derived. By providing explicit reasoning patterns, the \ac{cot} examples improve decision consistency, enhance interpretability, and reduce spurious or unstable reasoning behaviors across agents.
\begin{reasoning}[label={cot:wallet_agent}]{Wallet Agent.}
\begin{minipage}[t]{0.49\linewidth}
T-statistic of Returns: 24.39

Return Standard Deviation: 0.84

Last Return: 0.12

Ten-to-Six Return: 0.79

Time Since Last Trade: 371

\end{minipage}
\hfill
\begin{minipage}[t]{0.50\linewidth}
Average Return: 1.25

Number of Trades: 4114

Five-to-One Return: 0.14

Fifteen-to-Eleven Return: 0.48

Time Since First Trade: 19118974

\end{minipage}

\begin{lstlisting}[
  style=jsonplain,
  caption={Wallet Agent Reasoning (Positive Case)},
]
{"reasoning":{"Statistical Significance":"Gate check: require t-statistic > 1.645 (one-tailed 5%). Observed = 24.39. Since 24.39 > 1.645, the gate passes.","Profitability":"Check average return > 1.00. Observed = 1.25. Since 1.25 > 1.00, this check passes.","Risk Profile":"Check standard deviation < 1.00. Observed = 0.84. Since 0.84 < 1.00, this check passes.","Temporal Robustness":"Require 5-to-1, 10-to-6, and 15-to-11 returns all > 1.00. Observed = (0.12,0.14,0.79,0.48). All exceed 1.00, so this check passes.","Statistical Robustness":"Require number of trades > {25% Training Set}. Observed = 4114. Since 4114 > {25% Training Set}, this check passes.","Activity and Longevity":"Recency requires time since last trade < {75% Training Set}: 371 < {75% Training Set} passes. Longevity prefers time since first trade > {25% Training Set}: 19118974 > {25% Training Set}, so this check passes.","Summary":"All checks pass (significance, profitability, risk control, temporal robustness, longevity, and sample size). The wallet is attractive for next-trade copy trading."},"result":true}
\end{lstlisting}

\begin{minipage}[t]{0.49\linewidth}
T-statistic of Returns: 0.45

Return Standard Deviation: 0.31

Last Return: 0.02

Ten-to-Six Return: 0.11

Time Since Last Trade: 10

\end{minipage}
\hfill
\begin{minipage}[t]{0.50\linewidth}
Average Return: 0.00

Number of Trades: 461

Five-to-One Return: -0.02

Fifteen-to-Eleven Return: -0.12

Time Since First Trade: 15449
\end{minipage}
\begin{lstlisting}[
  style=jsonplain,
  caption={Wallet Agent Reasoning (Negative Case)},
]
{"reasoning":{"Statistical Significance":"Gate check: require t-statistic > 1.645 (one-tailed 5%). Observed = 0.45. Since 0.45 < 1.645, the gate fails.","Profitability":"Require average return > 0.00. Observed = 0.00. Since 0.00 is not greater than 0.00, this check fails.","Risk Profile":"Check standard deviation < 1.00. Observed = 0.31. This passes, but cannot compensate for failed core profitability and significance gates.","Temporal Robustness":"Require all horizons > 1.00. Observed = (0.02,-0.02,0.11,-0.12). Multiple horizons are below 0.00, so this check fails.","Statistical Robustness":"Require number of trades > {25% Training Set}. Observed = 461. Since 461 < {25% Training Set}, this check fails.","Activity and Longevity":"Recency 10 < {75% Training Set} passes; longevity 15449 < {25% Training Set} not met. Auxiliary signals cannot offset failed core checks.","Summary":"Core gates fail (statistical significance and profitability) and robustness is insufficient. The wallet is unsuitable for next-trade copy trading."},"result":false}
\end{lstlisting}
\end{reasoning}   
\begin{reasoning}[label={cot:coin_agent}]{Coin Agent.}

\begin{minipage}[t]{0.49\linewidth}
\textbf{Transaction Features:}

Bundle Bot: False

Sniper Bot: False

Bump Bot: True

Comment Bot: True

\textbf{Comment History:}

2025-01-17 15:06:36 -- 3yxCdw: Makers missing? Upgrade your strategy now!\\
...
\end{minipage}
\hfill
\begin{minipage}[t]{0.49\linewidth}
\centering
\textbf{Candlestick Chart:}

\vspace{0.3em}
\includegraphics[width=0.8\linewidth]{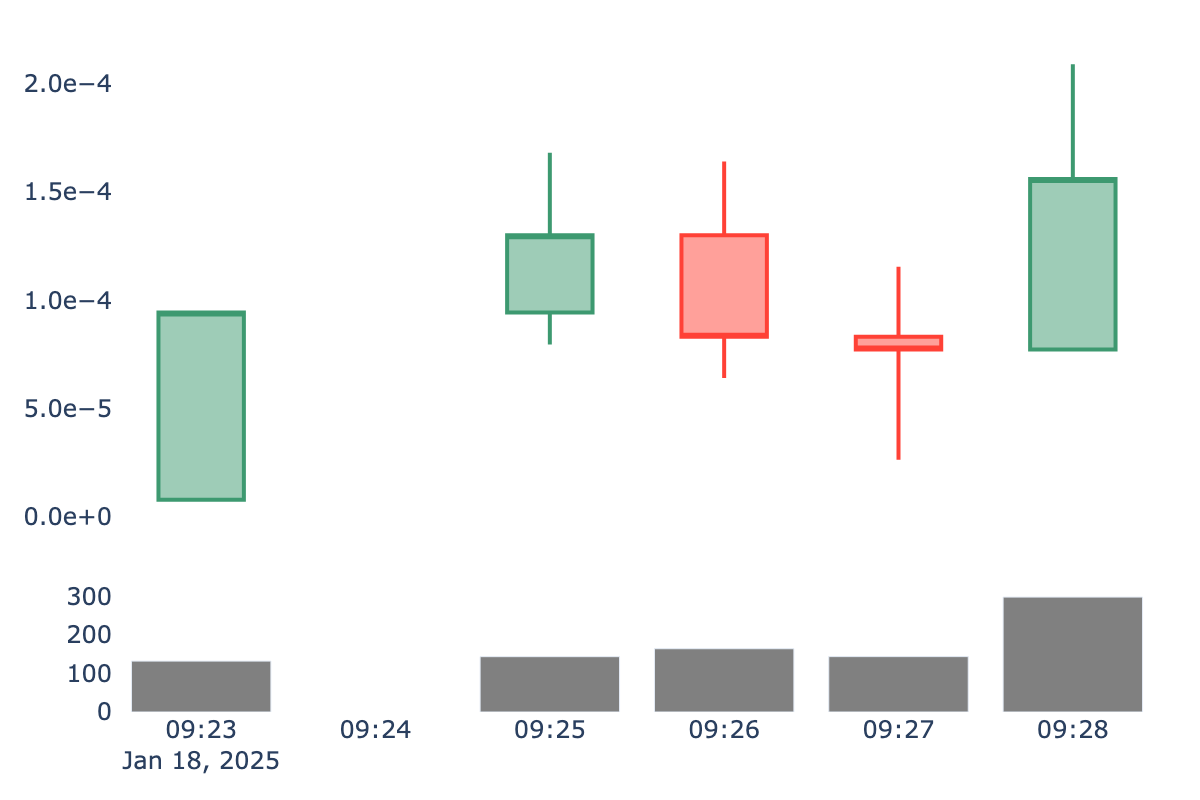}
\end{minipage}

\begin{lstlisting}[
  style=jsonplain,
  caption={Coin Agent Reasoning (Positive Case)},
]
{"reasoning":{"Bundle Bot":"Required check: Bundle Bot must be False. Observed = False. Since False == False, this check passes.","Candlestick Pattern":"Required check: chart should show gradual, sustained price discovery (no single-candle spike-and-dump). Observed pattern indicates a gradual increase, so this check passes.","Sniper Bot":"Auxiliary check: Sniper Bot ideally False. Observed = False. This reduces early predatory trading risk.","Bump Bot":"Auxiliary check: Bump Bot may be True as a weak visibility/support signal but is not required. Observed = True; treat as weakly supportive.","Comments":"Auxiliary check: discount repetitive spam; prefer evidence of non-boilerplate engagement. Observed Comment Bot = True; treat as caution, but not decisive if other required signals are clean.","Summary":"Required checks pass (no bundle and no pump-like candlestick signature). Auxiliary signals are not contradictory. Classify as a good investment opportunity."},"result":true}
\end{lstlisting}
\begin{minipage}[t]{0.49\linewidth}
\textbf{Transaction Features:}

Bundle Bot: True

Sniper Bot: True

Bump Bot: False

Comment Bot: False

\textbf{Comment History:}
\end{minipage}
\hfill
\begin{minipage}[t]{0.49\linewidth}
\centering
\textbf{Candlestick Chart:}
\includegraphics[width=0.8\linewidth]{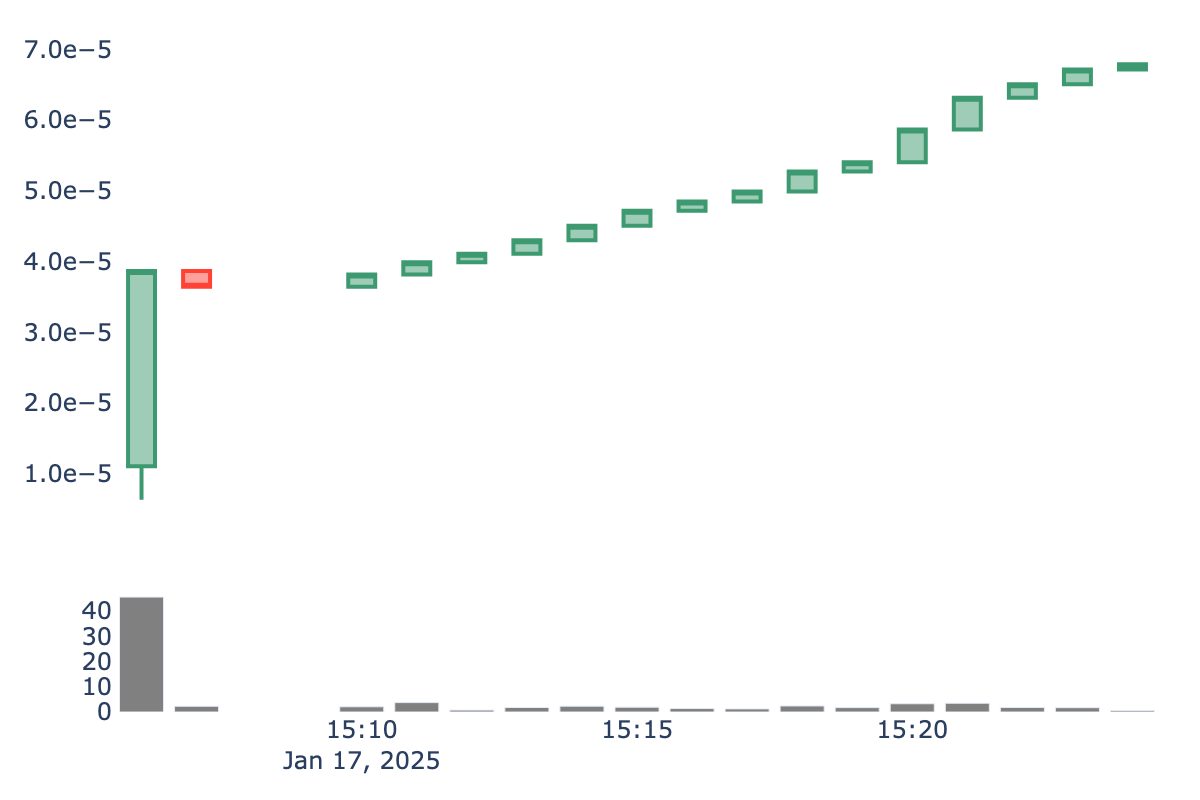}
\end{minipage}

\begin{lstlisting}[
  style=jsonplain,
  caption={Coin Agent Reasoning (Negative Case)},
]
{"reasoning":{"Bundle Bot":"Required check: Bundle Bot must be False. Observed = True. Since True != False, this check fails.","Candlestick Pattern":"Required check: chart should show gradual, sustained price discovery. Observed pattern resembles a spike/pump-like move, so this check fails.","Sniper Bot":"Auxiliary check: Sniper Bot ideally False. Observed = True; negative signal indicating elevated predatory early trading risk.","Bump Bot":"Auxiliary check: Bump Bot may be supportive if True. Observed = False; not supportive.","Comments":"Auxiliary check: prefer sustained, organic engagement. Observed Comment Bot = False; absent bot flags alone is not supportive without additional evidence, treat as non-decisive.","Summary":"One or more required checks fail (bundle present and pump-like price action). Auxiliary signals do not offset these failures. Classify as a poor investment opportunity."},"result":false}
\end{lstlisting}
\end{reasoning}
\begin{reasoning}[label={cot:timing_agent}]{Timing Agent.}

Time Since Launch: 3723

Trader Purchase Price: 1.19e-05

Trader Purchase Amount: 661.21

Trader Purchase Quantity: 55153573.25

\begin{lstlisting}[
  style=jsonplain,
  caption={Timing Agent Reasoning (Positive Case)},
]
{"reasoning":{"Time Since Launch":"Require time since launch < {75% Training Set}. Observed = 3723. Since 3723 < {75% Training Set}, this check passes.","Trader Purchase Price":"Require purchase price > {25% Training Set}. Observed = 1.19e-05. Since 1.19e-05 > {25% Training Set}, this check passes.","Trader Purchase Amount":"Require purchase amount > {25% Training Set}. Observed = 661.21. Since 661.21 > {25% Training Set}, this check passes.","Trader Purchase Quantity":"Require purchase quantity > {25% Training Set}. Observed = 55153573.25. Since 55153573.25 > {25% Training Set}, this check passes.","Summary":"All timing checks pass, indicating suitable entry timing for copy trading this wallet in the meme coin."},"result":true}

\end{lstlisting}

Time Since Launch: 85374

Trader Purchase Price: 4.95e-06

Trader Purchase Amount: 99.65

Trader Purchase Quantity: 6026170.61

\begin{lstlisting}[
  style=jsonplain,
  caption={Timing Agent Reasoning (Negative Case)},
]
{"reasoning":{"Time Since Launch":"Require time since launch < {75% Training Set}. Observed = 85374. Since 85374 > {75% Training Set}, this check fails.","Trader Purchase Price":"Require purchase price > {25% Training Set}. Observed = 4.95e-06. Since 4.95e-06 < {25% Training Set}, this check fails.","Trader Purchase Amount":"Require purchase amount > {25% Training Set}. Observed = 99.65. Since 99.65 < {25% Training Set}, this check fails.","Trader Purchase Quantity":"Require purchase quantity > {25% Training Set}. Observed = 6026170.61. Since 6026170.61 < {25% Training Set}, this check fails.","Summary":"One or more timing checks fail, indicating unsuitable entry timing for copy trading this wallet in the meme coin."},"result":false}
\end{lstlisting}
\end{reasoning}

\subsubsection{Prompt}
\label{apx:pmt}
We then present the prompts used by each agent in the proposed \ac{mas}, as provided in Prompt \autoref{pmt:wallet_agent}, \autoref{pmt:coin_agent}, and \autoref{pmt:timing_agent}. The prompts specify the agent’s role, available inputs, and decision objectives, and instruct the \ac{llm} to reason before producing a final decision. This design ensures consistent task alignment across agents while allowing each agent to specialize in its respective subtask.
\begin{prompt}[label={pmt:wallet_agent}]{Wallet Agent.}

You are a professional on-chain analyst specializing in meme coin wallet trading behavior. You will be given features of a wallet's historical meme coin trading activity. Your task is to assess whether the wallet's next trade will be profitable so that a user can copy trade from it.

\textcolor{blue}{\{\texttt{CoT \autoref{cot:wallet_agent}}\}}

T-statistic of Returns: \textcolor{blue}{\{\textit{$t$-stat}\}}

Average Return: \textcolor{blue}{\{\textit{Return (all)}\}}

Return Standard Deviation: \textcolor{blue}{\{\textit{Return Standard Deviation}\}}

Number of Trades: \textcolor{blue}{\{\textit{Number of Trades}\}}

Last Return \textcolor{blue}{\{\textit{Return (1st)}\}}

Five-to-One Return: \textcolor{blue}{\{\textit{Return (1st-5th)}\}}

Ten-to-Six Return: \textcolor{blue}{\{\textit{Return (6th-10th)}\}}

Fifteen-to-Eleven Return: \textcolor{blue}{\{\textit{Return (11th-15th)}\}}

Time Since Last Trade: \textcolor{blue}{\{\textit{Time Since Last Trade}\}}

Time Since First Trade: \textcolor{blue}{\{\textit{Time Since First Trade}\}}

\end{prompt}
\begin{prompt}[label={pmt:coin_agent}]{Coin Agent.}

You are a professional on-chain analyst specializing in meme coin investment potential. You will be given various transaction features, candlestick chart, and comment history related to a meme coin. Your task is to assess whether the meme coin is a good investment opportunity.

\textcolor{blue}{\{\texttt{CoT \autoref{cot:coin_agent}}\}}

\begin{minipage}[t]{0.49\linewidth}
\textbf{Transaction Features:}

Bundle Bot: \textcolor{blue}{\{\textit{Bundle Bot}\}}

Bump Bot: \textcolor{blue}{\{\textit{Bump Bot}\}}

\textbf{Comment History:} \textcolor{blue}{\{\textit{Comments}\}}

\end{minipage}
\hfill
\begin{minipage}[t]{0.50\linewidth}
\vspace{0.3\baselineskip}
Sniper Bot: \textcolor{blue}{\{\textit{Sniper Bot}\}}

Comment Bot: \textcolor{blue}{\{\textit{Comment Bot}\}}

\textbf{Candlestick Chart:} \textcolor{blue}{\{\textit{Candlestick}\}}
\end{minipage}

\end{prompt}
\begin{prompt}[label={pmt:timing_agent}]{Timing Agent.}

You are a professional on-chain analyst specializing in meme coin wallet trading behavior. You will be given features of a wallet's historical meme coin trading activity. Your task is to assess whether the wallet's next trade will be profitable so that a user can copy trade from it.

\textcolor{blue}{\{\texttt{CoT \autoref{cot:timing_agent}}\}}

Time Since Launch: \textcolor{blue}{\{\textit{Time Since Launch}\}}

Trader Purchase Price: \textcolor{blue}{\{\textit{Trader Purchase Price}\}}

Trader Purchase Amount: \textcolor{blue}{\{\textit{Trader Purchase Amount}\}}

Trader Purchase Quantity: \textcolor{blue}{\{\textit{Trader Purchase Quantity}\}}

\end{prompt}

\subsection{Smart Money Return, Copier Return, and Imitation Penalty}
\label{apx:leader_copier_return}
In this section, we provide proofs of \autoref{lemma:return} and \autoref{theorem:imitation_penalty}.


\begin{proof}[Proof of \autoref{lemma:return}]

Under the constant-product invariant $XY = k$ derived in \autoref{apx:bonding_curve_mechanism}, executing a trade of size $q$
at effective reserve $Y$ induces the following SOL flow:
{\small
\[
\Delta X(Y,q)
\;\equiv\;
\frac{k}{Y - q} - \frac{k}{Y}
\;=\;
\frac{kq}{Y(Y-q)}.
\]
For buys ($q>0$), $\Delta X(Y,q)>0$ represents SOL spent; for sells ($q<0$),
$\Delta X(Y,q)<0$ represents SOL received.
}
The smart money’s total SOL expenditure and proceeds across all trades are
{\small\[
X^{(S)}_{\mathrm{in}}
=
\sum_{t:q_t>0} \Delta X(Y^{S}_{t}, q_t),
\qquad
X^{(S)}_{\mathrm{out}}
=
-\sum_{t:q_t<0} \Delta X(Y^{S}_{t}, q_t),
\]
}
yielding the smart money’s net return
{\small\[
R^{(S)} \equiv \frac{X^{(S)}_{\mathrm{out}}}{X^{(S)}_{\mathrm{in}}} - 1 = \frac{-\sum_{t:q_{t}<0} \frac{kq_{t}}{Y^{S}_{t}(Y^{S}_{t}-q_{t})}}{\sum_{t:q_{t}>0} \frac{kq_{t}}{Y^{S}_{t}(Y^{S}_{t}-q_{t})}} - 1.
\]
}
A copier wallet replicates each trade immediately after the smart money.
Because the smart money’s trade updates the bonding curve state, the copier
faces a strictly worse reserve:
{\small\[
Y^{C}_{t} = Y^{S}_{t} - q_t.
\]
}
The copier’s total SOL expenditure and proceeds are therefore
{\small\[
X^{(C)}_{\mathrm{in}}
=
\sum_{t:q_t>0} \Delta X(Y^{S}_{t} - q_t, q_t),
\qquad
X^{(C)}_{\mathrm{out}}
=
-\sum_{t:q_t<0} \Delta X(Y^{S}_{t} - q_t, q_t),
\]
}
with net return
{\small\[
R^{(C)} \equiv \frac{X^{(C)}_{\mathrm{out}}}{X^{(C)}_{\mathrm{in}}} - 1 = \frac{-\sum_{t:q_{t}<0} \frac{kq_{t}}{(Y^{S}_{t} - q_{t})(Y^{S}_{t}-2q_{t})}}{\sum_{t:q_{t}>0} \frac{kq_{t}}{(Y^{S}_{t} - q_{t})(Y^{S}_{t}-2q_{t})}} - 1.
\]
}
\end{proof}
\begin{proof}[Proof of \autoref{theorem:imitation_penalty}]
Consider any buy trade of size $d>0$ executed at effective reserve $Y$.
Under the bonding curve, the required SOL outlay is
{\small\[
\Delta X(Y,d) = \frac{k d}{Y(Y-d)}.
\]
Since the copier executes strictly after the smart money, it faces the
updated reserve $Y-d$ and therefore pays
\[
\Delta X(Y-d,d) = \frac{k d}{(Y-d)(Y-2d)},
\]
}
which is well-defined for $Y>2d$. The relative execution cost is
{\small\[
\frac{\Delta X(Y-d,d)}{\Delta X(Y,d)}
=
\frac{Y}{Y-2d}
>
1,
\]
}
implying that the copier strictly overpays on every replicated buy.
By symmetry, for any sell trade ($d<0$), the copier receives strictly less SOL than the smart money. Aggregating across all trades yields
{\small\[
X^{(C)}_{\mathrm{in}} > X^{(S)}_{\mathrm{in}},
\qquad
X^{(C)}_{\mathrm{out}} < X^{(S)}_{\mathrm{out}}.
\]
}
Since the net return is defined as
$R \equiv X_{\mathrm{out}}/X_{\mathrm{in}} - 1$ and is strictly increasing
in the ratio $X_{\mathrm{out}}/X_{\mathrm{in}}$, it follows immediately that
{\small\[
R^{(C)} < R^{(S)}.
\]
}
Therefore, the imitation penalty can be defined as:
{\small\[
    \epsilon = R^{(S)} - R^{(C)} > 0.
\]
}
Economically, $\epsilon$ captures the structural execution disadvantage of copy trading on bonding curves.
\end{proof}

\end{document}